\PassOptionsToPackage{numbers, compress}{natbib}
\documentclass{article}

\usepackage[final]{neurips_2023}

\usepackage[utf8]{inputenc} % allow utf-8 input
\usepackage[T1]{fontenc}    % use 8-bit T1 fonts
\usepackage{url}            % simple URL typesetting
\usepackage{booktabs}       % professional-quality tables
\usepackage{amsfonts}       % blackboard math symbols
\usepackage{nicefrac}       % compact symbols for 1/2, etc.
\usepackage{microtype}      % microtypography
\usepackage{xcolor}         % colors

\usepackage{colortbl}

\usepackage{url}
\usepackage{niravstyle}
\usepackage{graphicx}
\usepackage{import}
\usepackage{multirow}
\usepackage{multicol}
\usepackage{rotating}

\usepackage{subcaption}

\usepackage{caption}

\DeclareCaptionType{equ}[][]

\npdecimalsign{.}

\usepackage{booktabs}

\usepackage{siunitx}
\usepackage{amssymb}
\usepackage{amsmath}

\usepackage{pgfplots}
\pgfplotsset{compat=newest}
\usetikzlibrary{pgfplots.statistics}
\usepgfplotslibrary{groupplots}

\newcommand{\aod}[1]{\textcolor{blue!50!black}{AOD:~~#1}}
\newcommand{\tsf}[1]{\textcolor{brown!50!black}{TSF:~~#1}}
\newcommand{\revision}[1]{\textcolor{magenta!50!black}{#1}}
\newcommand{\todo}[1]{\textcolor{red!75!black}{TODO:~~#1}}

\newcommand{\done}[1]{\textcolor{green!75!black}{DONE:~~#1}}
\newcommand{\discuss}[1]{\textcolor{cyan!25!black}{DOING:~~#1}}

\newcommand{\method}{\fsc{HiGGs}\xspace}

\newtheorem{definition}{Definition}

\renewcommand{\nsa}[1]{}
\renewcommand{\add}[1]{}
\renewcommand{\del}[1]{}
{}
\renewcommand{\tsf}[1]{}
\renewcommand{\aod}[1]{}
\renewcommand{\revision}[1]{}
\renewcommand{\todo}[1]{}
\renewcommand{\done}[1]{}
\renewcommand{\discuss}[1]{}
\usepackage{siunitx}
\usepackage{enumitem}
\usepackage[inline]{enumitem}

\usepackage[ruled,vlined,linesnumbered]{algorithm2e}

\title{Size Matters: Large Graph Generation with \method}

\author{%
  Alex O. Davies $^\dagger$, Nirav S. Ajmeri $^\dagger$, and Telmo M. Silva Filho $^\ddagger$\\
  $^\dagger$ School of Computer Science\\
  $^\ddagger$ School of Engineering Mathematics and Technology\\
  University of Bristol, UK\\
  \texttt{\{alexander.davies,nirav.ajmeri,telmo.silvafilho\}@bristol.ac.uk}\\
}

\begin{document}

\maketitle

\begin{abstract}
Large graphs are present in a variety of domains, including social networks, civil infrastructure, and the physical sciences to name a few.
Graph generation is similarly widespread, with applications in drug discovery, network analysis and synthetic datasets among others.
While GNN (Graph Neural Network) models have been applied in these domains their high in-memory costs restrict them to small graphs. 
Conversely less costly rule-based methods struggle to reproduce complex structures.
% GNN models for graph generation have been employed in each of these domains, but on datasets of small graphs, due to their high in-memory costs.
We propose \method (Hierarchical Generation of Graphs)
as a model-agnostic framework of producing large graphs with realistic local structures.
\method uses GNN models with conditional generation capabilities to sample graphs in hierarchies of resolution.
% \method samples graphs in hierarchies, and makes use of models with conditional generation capabilities.
As a result \method has the capacity to extend the scale of generated graphs from a given GNN model by quadratic order.
As a demonstration we implement \method using DiGress, a recent graph-diffusion model, including a novel edge-predictive-diffusion variant edge-DiGress.
We use this implementation to generate categorically attributed graphs with tens of thousands of nodes.
These \method generated graphs are far larger than any previously produced using GNNs.
% Graphs generated with \method are up to two orders of magnitude larger than those feasible from DiGress.
% MMD quality scores are competitive with graphs produced by other models of only a few hundred nodes.
Despite this jump in scale we demonstrate that the graphs produced by \method are, on the local scale, more realistic than those from the rule-based model BTER.

\end{abstract}

% \todo{
% \begin{enumerate}
%     \item Reformat contributions
%     \item Low/High Res Images
%     \item Large graph visualisations in supplemental
%     \item review if certain titles are needed
% \end{enumerate}
% }

\section{Introduction}
\label{sec:introduction}

Graphs have long been an area of interest for generative models.
Applications for generated graphs include novel drug synthesis \cite{Gilmer2017NeuralChemistry}, synthetic social networks \cite{Davies2022RealisticNetworks}, neural network wirings \cite{Xie2019ExploringRecognition} and in network science \cite{Watts1998CollectiveNetworks}. 
Current GNN graph generation methods have not been applied to graphs of more than a few thousand nodes, and the generation of graphs of this size with attribution is well beyond the current state-of-the-art. 
% The most efficient models (for example GRAN from \citet{Liao2019EfficientNetworks}) scale to a few thousand nodes without attribution.

We propose Hierarchial Generation of Graphs (\method) as a method of producing much larger graphs than can be generated using a single model. 
Making use of the modular structure inherent to many graph types, and relying on the ability of newer models such as DiGress \cite{Vignac2023DiGress:Generation} to conditionally generate attributed graphs, we break the generation process into hierarchies.
Each hierarchy takes a progressively lower-resolution view of the sampled graph.
The upper hierarchy conditions the generation of graphs in the lower, as well as the necessary edge-prediction between such lower-hierarchy graphs.
We demonstrate that our implementation using DiGress can produce categorical features for nodes, edges and whole graphs with tens of thousands of nodes.
This implementation includes a novel model edge-DiGress, which employs edge-predictive diffusion.
The core contributions of \method are: (i) \method is flexible in which models can be used in implementation, assuming some model capabilities such as conditional generation; (ii) \method has the capacity to extend the scale of graphs that can be generated by its component models by quadratic order, e.g. we extend the size of graphs produced using DiGress from around  \np{200} nodes to more than \np{20000} nodes; (iii) the \method framework allows the production of graphs with both node and edge attributes of a scale far larger than is possible with standalone GNN models; and (iv) \method can produce graphs of a scale comparable to rule-based methods while retaining realistic graph structures on the few-hundred-node scale.
% \begin{description}
%     \item[Model Agnostic Framework] \method is flexible in which models can be used in implementation, assuming some model capabilities such as conditional generation.
%     \item[Generated Graph Size] \method has the capacity to extend the scale of graphs that can be generated by its component models by quadratic order. In this implementation we extend the size of graphs produced using DiGress from around  \np{200} nodes to more than \np{20000} nodes.
%     \item[Node \& Edge Attribution] The \method framework allows the production of graphs with both node and edge attributes of a scale far larger than is possible with standalone GNN models.
%     \item[Realistic Local Structure] \method can produce graphs of a scale comparable to rule-based methods while retaining realistic graph structures on the few-hundred-node scale.
% \end{description}
We demonstrate the efficacy of \method on three datasets of increasing size.
% a dataset of SBM graphs from \citet{Martinkus2022SPECTRE:Generators}, then the much-used Cora dataset \cite{McCallum2000AutomatingLearning} at \np{2708} nodes and the MUSAE-Facebook page-page dataset \cite{Rozemberczki2021Multi-ScaleEmbedding} at \np{22470} nodes.
To our knowledge no other deep-learning method can produce graphs close to this scale.
% Through the application of \method to these datasets we investigate the following research questions:

% \subsection{Research Questions}

% \begin{enumerate}[label=\textbf{RQ\arabic*},itemindent=*,]
    % \item How can we exploit and encode meaningful hierarchies from graphs?

\tsf{Shouldn't the novel edge-DiGress be listed as a contribution?}
\aod{Added above - not a core contribution but is now mentioned}
\tsf{Why 3000 for the large threshold?}
\aod{Changed to 2000 - smallest graph with plural thousands of nodes}

We evaluate the ability of our implementation of \method in producing graphs, and through varying graph domains and feature augmentation, investigate its limitations in reproducing distinct graph characteristics at both whole-graph and local scales.
    % \nsa{say \fsl{large} instead of larger?}  \nsa{In the follow up description, define \fsl{large} as graphs with V > \np{2000}. Cite something for large}
    % \item How can \method allow production of larger graphs?
We implement \method using DiGress for higher and lower hierarchies and a novel edge-predictive implementation edge-DiGress. 
% We vary the conditioning attributes in the higher hierarchy to investigate how a user might optimise \method for different target graph characterstics.
% \citet{Erciyes2021LargeAnalysis} describe large graphs as \textit{``thousands of vertices and tens of thousands of edges between these vertices''}. 
We set the threshold for a graph to be ``large'' at \np{2000} nodes, but produce graphs an order of magnitude larger than this.
    % \item What are the limitations of \method?
% Through the evaluation of DiGress and edge-DiGress at each sampling stage, and of \method in sampling full large graphs, we are able to explore the limitations of this implementation.
% We also include discussion of which of these limitations could be alleviated, and which are inherent to the \method framework.
% \end{enumerate}

% The rest of the paper is structured as follows. 
% Section~\ref{sec:related} describes the current state of graph generation methods. 
% Section~\ref{sec:methodology} details \method as a new framework, including details of the implementation used in this work, and a discussion of its time and memory complexities. 
% Section~\ref{sec:experiments} describes our experiments and results. 
% Section~\ref{sec:discussion} discusses our findings and lists the assumptions we've made, and the threats to the validity of this work.
% Section~\ref{sec:conclusion} presents our conclusions. 

\section{Background and Related Work}
\label{sec:related}

Our work is closely related to graph generation, which is used in domains with greatly varied graph characteristics, and broadly splits into rule-based and GNN methods.
\citet{Zhao2020AGeneration} give an overview of graph generation in general.
Two approaches of Graph Neural Network (GNN) models have been developed for graph generation.
The first is one-shot methods, which output a whole graph at once.
These can use the whole graph structure, but often suffer higher complexities in both time and in-memory, as many such models scale $O(N^{>2})$ \cite{Bojchevski2018NetGAN:Walks, Simonovsky2018GraphVAE:Autoencoders, Wang2017GraphGAN:Nets}.
The second school, auto-regressive models, add new nodes or motifs progressively to construct graphs \cite{Bojchevski2018NetGAN:Walks, Liao2019EfficientNetworks}.
These have the opposite compromise to one-shot methods, in that they have lower complexities, but must go to greater lengths to include structural information. 
% Deep-learning approaches to graph generation vary considerably.
% Some approaches model graph construction as biased random walks, such as NetGAN \citet{Bojchevski2018NetGAN:Walks}.
% Others are ``one-shot'' methods, for example GraphGAN \cite{Wang2017GraphGAN:Nets}.
% Often models build on architectures from other forms of data with papers such as  \citet{DeCao2018MolGAN:Graphs} and \citet{Simonovsky2018GraphVAE:Autoencoders} adapting GANs and VAEs.
% These models are limited to small graphs due to issues presented with node orderings and the rapidly scaling memory requirements of graphs with increasing size. 
Recent models such as GRAN \citep{Liao2019EfficientNetworks} and TG-GAN \citep{Zhang2021TG-GAN:Constraints} are able to generate graphs of reasonable size (up to \np{2700} nodes). 
% \citet{Davies2022RealisticNetworks} show that GRAN is able to realistically reproduce social graphs at a lower scale, roughly \np{500} nodes. 
These efficient models generally deal only with un-attributed topologies, and models that can produce attributes still retain high complexity. 

% Of particular utility is \textit{conditional} generation - that is, generation of graphs given some target characteristic, for example volatility in molecule generation \cite{DeCao2018MolGAN:Graphs}.
% Models with this capability are again more expensive. 
The recent development of diffusion models \cite{Ho2020DenoisingModels} and MPNN models \cite{Gilmer2017NeuralChemistry} has allowed development of, in the last year, several denoising diffusion models for graph structures. These are in their early stages and retain high memory and time complexities but have promising capacity for conditional generation. In the supplemental material we give details of a subset of GNN graph generation models.
\citet{Zhu2022AApplications} and \citet{Zhang2022DeepSurvey} provide detailed reviews using GNNs for graph generation, including example applications.
% \subsection{Rule-based Methods for Graph Generation}
Rule-based methods are often meant as an exploration of a mathematical model rather than as a method for generating graphs \cite{Chakrabarti2004R-MAT:Mining,Erdos1960ONGRAPHS}.
As such they often take some given parameter, for example, a connection probability matrix or a degree distribution \cite{Chakrabarti2004R-MAT:Mining,Erdos1960ONGRAPHS,Seshadhri2012CommunityGraphs}, and attempt to reconstruct the original graph as well as possible using that parameter.
This differentiates them from deep-learning methods, which might instead take a target characteristic, instead of an entire distribution.
Rule-based methods are, as a function of this simplicity, generally far more efficient than GNN methods, but can struggle to reproduce more complex graph features.
\method aims to combine the realism of GNN-generated graphs with the scale possible through rule-based methods.

% Human graphs often have the ``small world'' characteristic, where the average distance between nodes is small and consistent across graphs sizes.
% \citet{Watts1998CollectiveNetworks} propose a random generator that exhibits these desired properties: high clustering coefficients and small diameters.
% Each node has a set number of neighbours and connections between nodes are randomly sampled.
% % 
% This presents issues in that each node retains the same number of neighbours resulting in graphs that are over-simple or non-realistic.

% Real-world graphs present a power-law distribution in node degrees.
% A more recent generative method, R-MAT \cite{Chakrabarti2004R-MAT:Mining}, generates via a recursive division of the original graph's adjacency matrix, and results in graphs with this power-law distribution.
% % 
% R-MAT remains commonly used in recent works, often adapted with iterative changes \cite{Nettleton2016AGraphs,Gursoy2019AGeneration,Nettleton2021MEDICI:Generator}\nsa{List alphabetically}.
% This includes the distribution of node-level attributes across topologies generated by R-MAT.
% \citet{Ali2014SyntheticData} use \citet{Wang2011LeveragingSystems}'s generator for topology, then autonomously distribute node-level attributes according to the local topology of those nodes.
% Notably all of these methods often miss complex dynamics of social graphs, in particular in generating meaningful large-scale structure.

\section{Schematics}
\label{sec:methodology}

% Here we detail \method, the implementation we use to perform our experiments, and discuss the complexity of this implementation.

% \subsection{\method}

\method produces graphs through hierarchical sampling. 
First a low-resolution version of the final graph is sampled.
Next, conditioned on that low-resolution graph, a set of high-resolution component graphs are sampled and joined.
This is analogous to patch-based image super resolution \cite{Scupakova2019ASpectrometry}, but with the necessary additional step of inter-high-resolution edge-sampling.
Previous motif-based \cite{Jin2020HierarchicalMotifs} or block-generative models \cite{Liao2019EfficientNetworks} add nodes or motifs sequentially, and use only one model, whereas \method uses a separate model for each stage shown in the supplemental material.
% Figure~\ref{fig:process_diagram_pict} shows a  schematic of this process, and here we provide detail on each stage.

% \begin{figure*}[ht]
%     \centering
%     \includegraphics[width=\linewidth]{images/\method_Adjacency.drawio.png}
%     \caption{A basic schematic of the \method method. First an $h_2$ graph is sampled, which represents $h_1$ graph types and \textit{``are they connected''}-level inter-$h_1$ edges. Secondly each $h_1$ graph is sampled, conditioned on its type in the preceeding $h_1$ graph. Lastly, for each edge in $h_2$, the edges between the corresponding $h_1$ graphs are sampled.}
%     \label{fig:process_diagram}
% \end{figure*}

% \subsubsection{Terminology \& Notation}

% Here we formalise the notation we use in \method.

\begin{definition}[Hierarchy-1; $h_1$]
    The most granular, or highest-resolution, partition of the graph. Nodes are the original entities (users in social graphs).
    % In the original data each $h_1$ graph is a community found currently by Louvain detection.
\end{definition}

% \begin{definition}[Hierarchy-1$_x$ and Hierarchy-1$_y$; $h_{1x}, h_{1y}$]
%     Pairs of $h_1$ graphs that are known to be connected. $h_{1x}$ is un-connected (i.e. two separate component $h_1$ graphs) and $h_{1y}$ is with their inter-$h_1$ edges. 
%     % The ``glueing'' model ICE-DGD (Inter-Community Edge Discrete Graph Diffusion) predicts the new edges, $P(E_{1y}|\{E_{1x}\})$.
% \end{definition}

\begin{definition}[Hierarchy-2; $h_2$]
    Graph consisting of community-community links, or in this terminology, $h_1 \leftrightarrow h_1$ graphs. 
    Node and edge attributes in $h_2$ indicate some characteristic of $h_1$ graphs and how they inter-connect.
    % Currently each edge in a $h_2$ graph just shows whether in the original graph these communities are detected, but this might be extended to represent the extent to which they are connected. 
    % Classes for $h_2$ nodes are currently inhereted as the majority class of the nodes in the $h_1$ graph it represents.
\end{definition}

% \subsubsection{Stage One: $h_2$}

In Stage One the template $h_2$ graph is sampled. $h_2$ graphs are a low-resolution representation of larger graphs. Each node in $h_2$ represents a subgraph, with attributes describing that subgraph's characteristics.
% The type of an $h_1$ node in a graph is discrete, and an $h_2$ node cannot belong to more than one category.
As an example the category might represent the majority class of the $h_1$ graph that the node represents.
% Here the category of a node in $h_2$ is used to conditionally sample the $h_1$ graph it represents in Stage Two.
Edges in the $h_2$ graph $e = \{ h_{1,i}, h_{1, j} \}$ describe whether the two $h_1$ graphs present are connected. 
Edges in $h_2$ can be attributed, in which case the category of that edge is used as conditioning for the edge-prediction in Stage Three.

% \subsubsection{Stage Two: $h_1$}

% \aod{Some of this could be supplemental? Want this not to be specific to this implementation}

In Stage Two an $h_1$ graph is sampled for each node in the preceeding $h_2$ graph, with sampling conditioned on the attributes of that node.
Conditioning attributes here might be as simple as the majority class of the sub-graph.
However, as future models develop the ability to produce more complex node attributes, conditioning could encode topological features.
% How to attribute or categorise $h_1$ graphs is an open problem. Encoding node attributes might ensure a closer distribution to the real data in those attributes, and the same for edges, whereas encoding a topological metric could be used to optimise for topological features.
% \begin{itemize}
%     \item \textbf{Encode node classes} Label each $h_1$ graph according to some aggregation of node classes within that graph. For example $h_1$ graphs could be categorised according to the majority node category.
%     \item \textbf{Encode edge classes} By the same logic as encoding node class, $h_1$ graphs could be categorised according to the categories of edges they contain.
%     \item \textbf{Encode topological information} Labelling could be done according to a topological measurement of the graph, for example number of nodes or number of edges, though most schemas would require binning to reach a discrete and exclusive category. 
% \end{itemize}
% Which of these to use would require consideration of the most important thing to be reproduced.
% \subsubsection{Stage Three: $h_{1,i} \leftrightarrow h_{1, j}$}
In Stage Three, for each edge in the initial $h_2$ graph $e=\{ h_{1,i}, h_{1,j} \}$, the edges between the $h_{1,i}$ and $h_{1,j}$ are sampled, $h_{1,i} \leftrightarrow h_{1, j}$. 
% The edge $e$ here might itself be attributed, which could then be used as conditioning for this inter-$h_1$ edge sampling. 
Crucially, and unlike block-generative or motif-based models, stages two and three consist of many independent jobs, making them trivially paralellisable.
% As in choosing a categorisation scheme for $h_1$ graphs, the categories for edges in $h_2$ that are used to condition this edge-predictive sampling can either emphasise a distribution of attributes or a topological feature. 

% For example, if better re-produced topology is more desireable, edge classes in $h_2$ might describe the number of edges between the $h_1$ graphs in question.
% Alternatively, should a given distribution of edge categories be more important, an edge in $h_2$ could be classified according to some binning of that distribution.

% 

% \subsubsection{$h_2$ Sampling}

\subsection{Implementation}

The \method framework requires several component models and algorithms.
For training dataset construction a segmentation algorithm is required.
This should produce subgraph partitions that represent samples from the same distributions (i.e. communities in a social network).
If the original data is one or few large graphs, then it should have a random element ensuring that on successive applications the same partitions are not produced. 
Individual partitions are then nodes in $h_2$ training samples, each partition itself an $h_1$ training sample, and each pair of connected partitions a training sample for edge-sampling.

% \begin{table}[ht]
%     \centering
%     \begin{tabular}{c|c|c|c}
%     Segmentation   &  Stage One ($h_2$)  &  Stage Two ($h_1$)  &  Stage Three ($h_{1,i} \leftrightarrow h_{1, j}$) \\\midrule
%     Louvain        &  DiGress    &  DiGress    &  edge-DiGress (Section~\ref{sec:edge-DiGress}) 
%     \end{tabular}
%     \caption{Component models for this implementation of DiGress.}
%     \label{tab:implementation}
% \end{table}

Three graph-generative models are then required for stages one, two and three, one of which is edge predictive.
The remaining two models are trained to produce $h_1$ and $h_2$ graphs. 
We implement, as a baseline demonstration, \method using an adaption of the Discrete Denoising Diffusion (DiGress) model from \citet{Vignac2023DiGress:Generation}.
This model handles discrete graph, node and edge classes, and applies discrete noise during the forward diffusion process.
% This is analogous to diffusion for categorisation with the absence of an edge treated as its own class. 
For this implementation we use repeat applications of Louvain partitioning \cite{DeMeo2011GeneralizedNetworks} to construct training sets of $h_1$, $h_2$ and $h_1 \leftrightarrow h_1$ graphs.
Louvain segmentation is a modularity-optimising algorithm for dividing graphs into `communities', beginning with individual nodes and iteratively merging them until a set of optimised partitions is reached.
% The size of these partitions is broadly controlled by the `resolution' parameter.
As in \citet{Vignac2023DiGress:Generation}, we compute extra features at each diffusion timestep. We implement calculation of clustering and an approximation for diameter, detailed in the supplemental material, but only use clustering as an extra feature in our experiments.\footnote{All code and data available here: \url{https://github.com/higgs-neurips-23/HiGGs}}

% \subsubsection{edge-DiGress} 
% \label{sec:edge-DiGress}

We implement edge predictive diffusion as an extension of DiGress. At each stage of the DiGress noise application and sampling process, we pass a mask of whether each node-node pair are in the same $h_1$ graph, and apply noise only on edges where they are not. This means that during training edge-DiGress learns to remove noise from edges between, but not within, two $h_1$ graphs.
The advantage of edge-DiGress over other edge prediction models is that it can be conditioned, and samples all required edges simultaneously.
% For example, if the conditioning edge in $h_2$ indicates a high density of edges, edge-DiGress should produce a number of inter-$h_1$ above the average in the training data.
Further  details, including complexities, can be found in the supplemental material.

\section{Experiments}
\label{sec:experiments}

% Experiments aim to evaluate the ability of our \method implementation, first in comparison to DiGress on a dataset within its graph size limits, then on graphs beyond the limitations of current models.

% \subsection{Data}

% \begin{table}[!htb]
% \caption{Details of our large graph datasets
% \nsa{formatted differently compared to Table 1; add bottomrule here; Table 1 could be formatted using toprule, midrule and bottomrule}}
% \label{tab:datasets}

% \centering
%     % \begin{tabular}{c||c|c|c|c}
%     \begin{tabular}{l n{5}{0} n{5}{0} n{5}{0} n{3}{0} n{1}{7} n{1}{4}}
%     \toprule
%     &   {Nodes}         &   {Edges}     &   {Diameter}    &   {Clustering}   &   {Density}    &    {Transitivity} \\\midrule
%     Cora        &   2810          &   7981      &   17          & 0.277         &   0.00202    &    0.114   \\
%     Facebook    &   22470         &   171002    &   15          & 0.360         &   0.000680   &    0.232   \\
%     Iceland     &   27980         &   33190     &   246         & 0.0472        &   0.0000847  &    0.0732  \\
% \end{tabular}
% \end{table}

% \subsubsection{Stochastic Block Model}

We employ three graph datasets in our experimentation, two of which are well beyond the scale limitations of current models. 
As in \citet{Vignac2023DiGress:Generation}, we first evaluate the performance of \method in reproducing the Stochastic Block Model (SBM) benchmark graphs from \citet{Martinkus2022SPECTRE:Generators}.
% These are \np{200} graphs of between \np{2} and \np{5} communities, each of which is between \np{20} and \np{40} nodes, giving a graph size range of \np{40} to \np{200} nodes.
% The inter-community and intra-community edge probabilities are \np{0.05} and \np{0.3} respectively. 
This dataset allows comparison of \method against simply using one of its component models at these low graph sizes.
% \subsubsection{Citation Graph (Cora)}
As a smaller large graph we use the popular Cora citation network \cite{Sen2008CollectiveData}.
% Each node represents an academic article or publication, categorised into 7 different fields of machine learning.
% Though at the lower limit of where we define large graphs with \np{2810} nodes i
% It is commonly used in graph generation applications \cite{Bojchevski2018NetGAN:Walks, Fan2019LabeledNetworks}, though at \np{2810} nodes other works aim to reproduce sampled sub-graphs of the whole network, in the range $150 \leq |V| \leq 200$.
% As a citation network the Cora graph is directed, we take the upper part of the adjacency matrix to produce an un-directed graph, using only the largest connected component.
As an example of a large social graph we use the Facebook Page-Page graph published by \citet{Rozemberczki2021Multi-ScaleEmbedding}.
This is an undirected graph of $|V|$ = \np{22470} nodes.
% Nodes are categorised into the type of Facebook page. 
% The density $\rho$ of this graph is comparatively high, and at a diameter of 15 edges, produces densely connected $h_2$ graphs when Louvain community detection is applied.
% which is what we use to segment this graph into hierarchies.
% \subsubsection{Road Graph (Iceland)} 
% We use the Python package OSMNX (Open Street Map graphX) from \citet{Boeing2017OSMnx:Networks} to collect the entire (driving) road graph available for Iceland from Open-Street Maps. This produces a graph of \np{27980} nodes, with each node a junction, and each edge a road between junctions. Road networks are geometric graphs, meaning that they derived from points on some plane, here with the surface of Iceland the plane in question. This graph is very sparse $\rho \sim 8 \times 10^{-5}$, and as road graphs are near-planar, has far sparser $h_1$ and $h_2$ graphs under Louvain segmentation.
% \subsection{Training and Sampling}
% Model training and sampling is conducted on a workstation computer equipped with an NVIDIA A6000 48gb GPU. For each dataset we fit the model for each stage with parameters based on the nature of the large graph.
The most critical parameter for \method is the resolution $r$ used by Louvain partitioning.
% As resolution increases so does the number of partitions and there is a corresponding decrease in the size of those partitions.
If $r$ is too high then the number of edges in $h_2$, and the level of information that must be encoded in $h_2$,  increases rapidly.
Conversely, if $r$ is too low, then the size of $h_1$ graphs can be too large for pairs to fit in-memory during edge-sampling.

% Social networks famously exhibit small world properties \cite{Watts1998CollectiveNetworks}, meaning consistently small diameters over a range of graph sizes. In contrast geometric graphs, such as road networks, generally have diameters that increase with graph size. \citet{Loukas2020WhatWidth} demonstrate that in order to encode (and therefore learn to generate) the diameter of a graph, a GNN must have a number of layers approximately equal to that diameter.

% \subsubsection{Conditioning Regimes}

% Our large graph datasets represent very different underlying distributions.
% Further, the Cora and Facebook graphs include node categories, but not edge categories, and for the Icelandic road graph the opposite is true.
% To this end different conditioning regimes should be used in model training and graph sampling.

For the two node-labelled datasets, Cora and Facebook, we categorise $h_1$ graphs by their majority node category.
% The sampling of $h_1$ graphs is conditioned on that node class from $h_2$.
% For simplicity we do not condition edges in $h_2$, assuming that the number of edges between $h_1$ graphs is a distribution that can be adequately learnt by our edge-sampling model. 
To ensure that pairs of $h_1$ graphs can fit in-memory for edge-prediction (Stage Three) we use a high resolution $r=10$ in Louvain segmentation for the Facebook dataset.
We are able to use a more conservative $r=2$ for the SBM and Cora datasets.
% We do not use edge categorisation in $h_2$ and hence do not use conditional edge-sampling in Stage Three.
% The version of DiGress used in this \method implementation is from the pre-print version of \citet{Vignac2023DiGress:Generation}, and so does not use their updated regressive conditioning method.
For conditional sampling we simply fix the category of the graph during diffusion. % ($y$ in \citet{Vignac2023DiGress:Generation}'s terminology).
% Superior performance might be achieved if this updated method were used, but as t
This work only provides a baseline demonstration of the \method framework and we leave improved conditioning methods as an area for future study.
The original DiGress work provides an ablation study \cite{Vignac2023DiGress:Generation}.

% For the Icelandic road network we similarly condition $h_1$ graphs (and $h_2$ nodes) by majority road (edge) class. Roads (edges) are simply the type of road, ie primary, secondary, tertiary and residential. Where our data has missing data for road type we assume that roads are residential. As for Facebook and Cora we do not use edge classes in $h_2$ and as such do not condition edge-sampling.

\subsection{Benchmarking}

As a benchmark we compare the performance of this \method implementation against DiGress to establish how much performance is lost by using \method at graph scales within current model limitations.
In contrast to the Facebook and Cora datasets, where we construct training data by repeated application of Louvain partitioning, for the SBM dataset we apply Louvain detection once on each separate graph to construct training sets.

Selecting a good benchmark model against which to compare the performance of this \method implementation is challenging on large graphs, as to our knowledge there are no existing GNN models that extend to this scale, and few recent rule-based models with the capacity to produce even categorical node attributes.
To this end we forgo using a model that produces attributes and use the Block Two-Level Erdos-Renyi (BTER) model from \citet{Seshadhri2012CommunityGraphs}.
BTER, like \method, assembles graphs as connected communities, and was specifically proposed to target the high clustering coefficients expected in real-world graphs like social networks.
% It targets a (given) specific degree distribution, guaranteeing equal numbers of nodes in sampled graphs, and in theory also ensuring a close-to-real degree distribution.
% However, without further assumptions or parameter setting, it cannot produce graphs scales different to those it is fit on.

\subsection{Results}

\begin{figure}[h]
     \centering
      \begin{subfigure}[b]{0.245\textwidth}
         \centering
         \includegraphics[width=\textwidth]{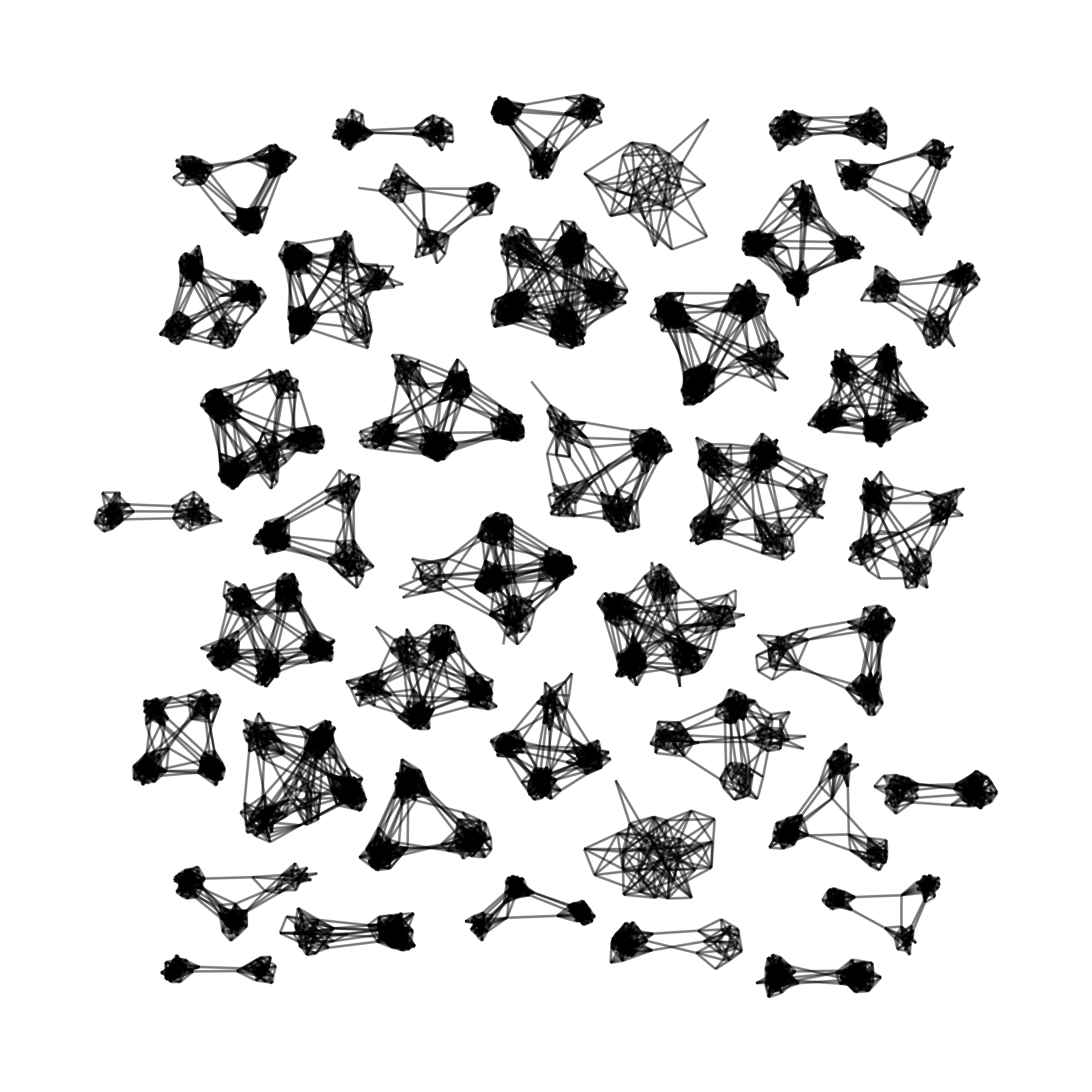}
         \caption{Original}
         \label{fig:sbm_real}
     \end{subfigure}
     \hfill
     \begin{subfigure}[b]{0.245\textwidth}
         \centering
         \includegraphics[width=\textwidth]{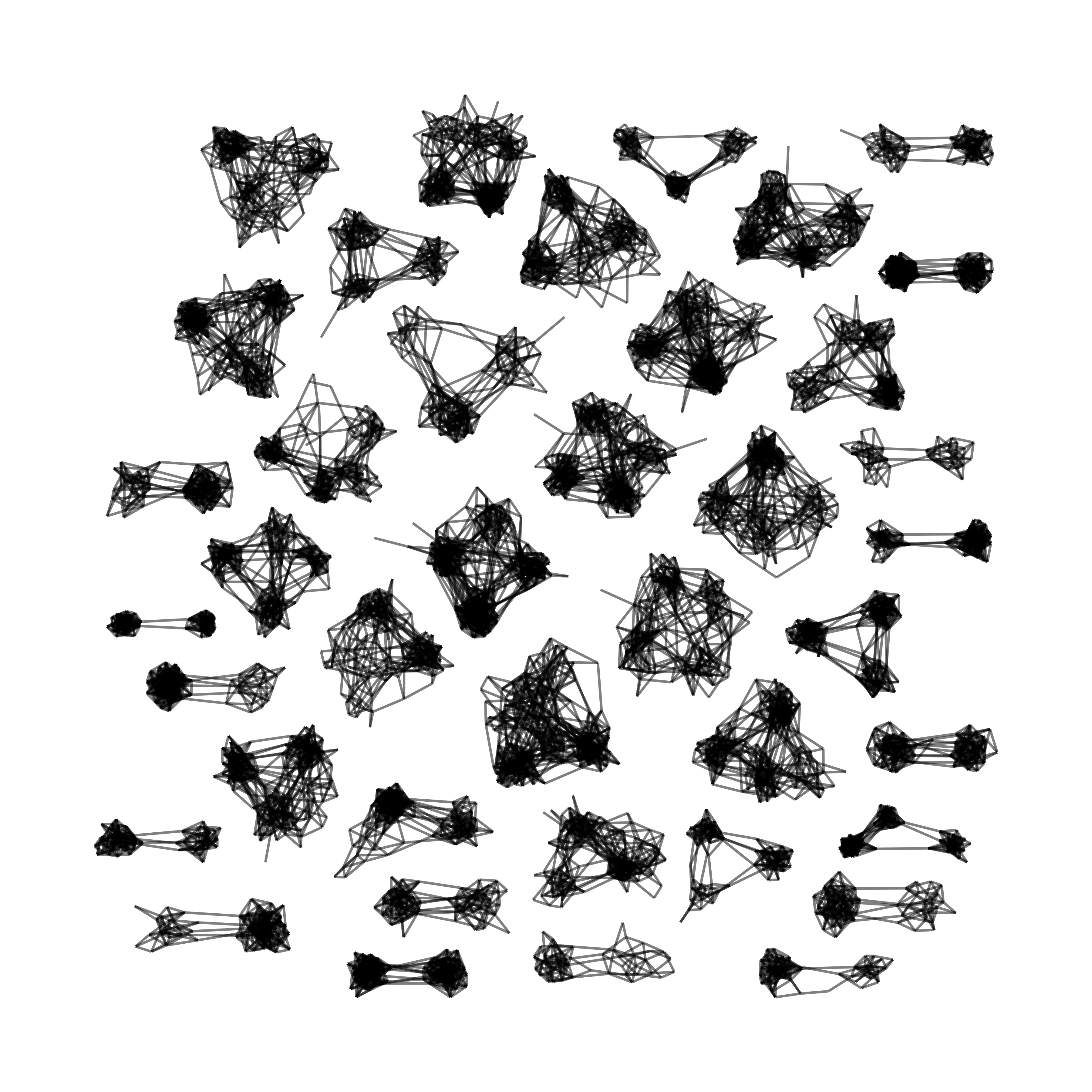}
         \caption{DiGress}
         \label{fig:sbm_dgd}
     \end{subfigure}
     \hfill
     \begin{subfigure}[b]{0.245\textwidth}
         \centering
         \includegraphics[width=\textwidth]{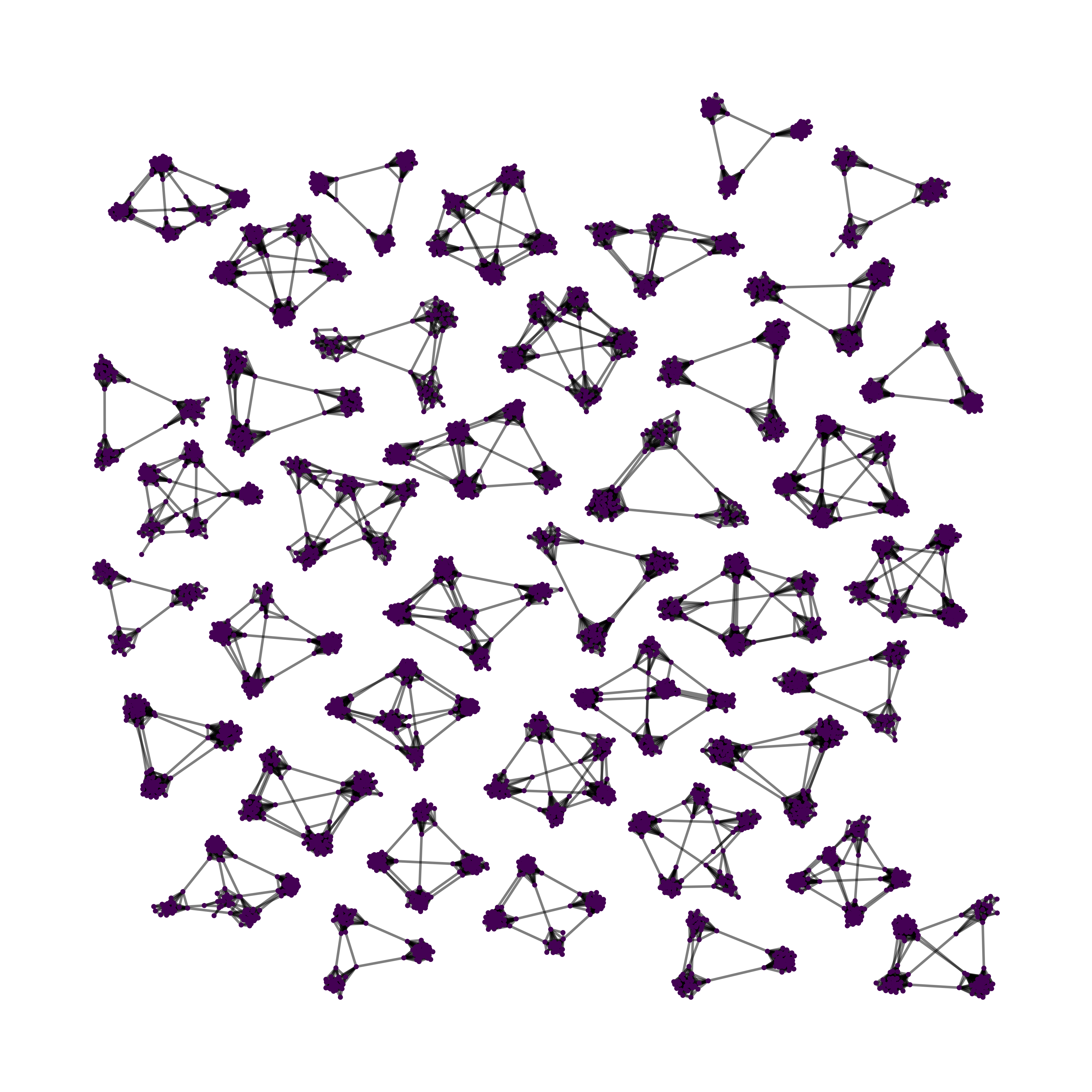}
         \caption{HiGGs}
         \label{fig:sbm_higgs}
     \end{subfigure}
     \hfill
     \begin{subfigure}[b]{0.245\textwidth}
         \centering
         \includegraphics[width=\textwidth]{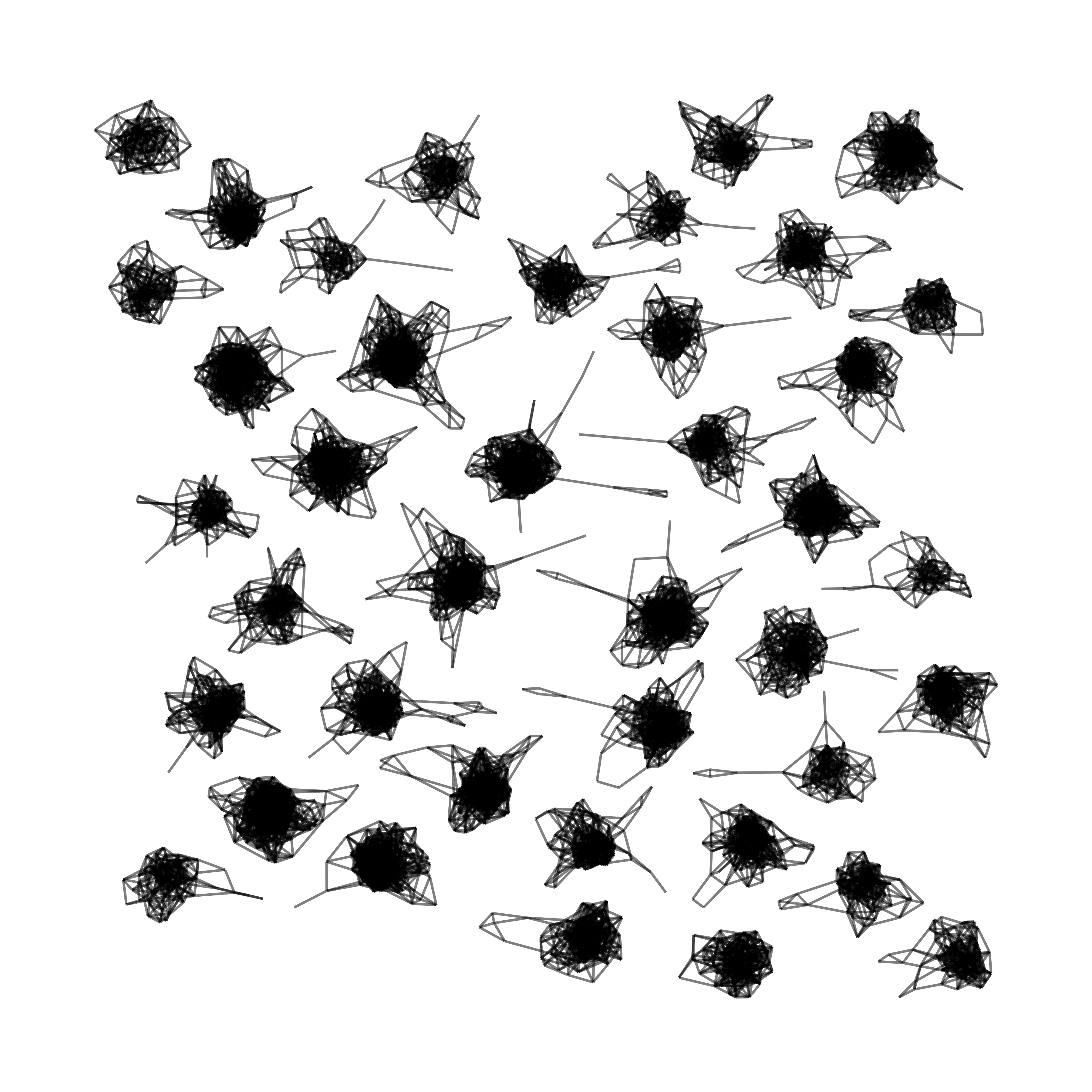}
         \caption{BTER}
         \label{fig:sbm_bter}
     \end{subfigure}
        \caption{Graphs from the SBM dataset with sampled counterparts from DiGress, \method and BTER. Layouts are through SFDP and GraphViz.}
        \label{fig:sbm-graphs}
\end{figure}

For the SBM dataset we sample 40 graphs (the size of the prescribed test-set) using DiGress, BTER and \method.
BTER graphs are fit on each individual test-set graph.
For the Cora and Facebook Page-Page datasets we use \method and BTER to sample one large graph.
Where multiple connected components are produced we take only the largest.
Aggregate results can be found in the supplemental material.
$N_C$ denotes the number of communities found under Louvain segmentation, and $|C|_{med}$ the median size of those communities.
Maximum Mean Discrepancy (MMD) scores can be found in Table~\ref{tab:mmd_results}.
Details of the measured quantities used for these scores is found in the supplemental material.
We exclude values where sample sizes of measured quantities are small.
% , namely aggregate community measurements on the SBM dataset, and number of nodes and spectral eigenvalues (Spectre) in the large sampled graphs for MMD.
Quantile-Quantile (QQ) plots and visualisations of sampling results can be found in the supplemental material for the Facebook graph and in the supplemental material for the Cora graph and SBM dataset.

\begin{table}[!htb]
\caption{MMD scores for models applied to each dataset. ``Communities'' denotes MMD calculated on communities re-sampled with Louvain partitioning.}
\label{tab:mmd_results}

\centering
    % \begin{tabular}{c||c|c|c|c}
    \begin{tabular}{l l n{1}{3} n{1}{4} n{1}{4} n{1}{5} n{1}{3}}
    \toprule
                                    &              &   {Nodes} &   {Degree}    &   {Clustering}    &   {Eccentricity}    &   {Spectre} \\\midrule
    \multirow{3}{*}{SBM}            & \method      &   0.547   &   0.286       &   0.411           &   0.0401            &   0.314     \\
                                    & DiGress      &   0.407   &   0.267       &   0.168           &   0.0193            &   0.195     \\
                                    & BTER         &   0.088   &   0.189       &   1.07            &   0.051             &   0.311     \\\midrule
    % ====================================================================================================================
    \multirow{2}{*}{Cora}           & \method      &  {--}     &   0.254       &   0.958           &   0.273             &   {--}      \\
                                    & BTER         &  {--}     &   0.439       &   0.281           &   0.00364           &   {--}      \\\midrule
    % ====================================================================================================================
    \multirow{2}{*}{Cora Communities}     & \method      &   0.662   &   0.069       &   0.451           &   0.0135            &   0.100     \\
                                    & BTER         &   0.598   &   0.290       &   0.417           &   0.00942           &   0.100     \\\midrule
    % ====================================================================================================================
    \multirow{2}{*}{Facebook}       & \method      &  {--}     &   2.00        &   0.904           &   0.229             &  {--}       \\
                                    & BTER         &  {--}     &   1.31        &   0.402           &   0.368             &  {--}      \\\midrule
    % ====================================================================================================================
    \multirow{2}{*}{Facebook Communities} & \method      &   0.744   &   0.0436      &   0.0918          &   0.00638           &   0.045     \\
                                    & BTER         &   0.620   &   0.104       &   0.481           &   0.0139            &   0.108     \\
    % Iceland   &   27980         &   33190     &   246         & 0.0472        &   0.0000847  &    0.0732  \\
    % Sampled   &   24651         &   29847     &   76          & 0.0125        &   0.0000982  &    0.0171  \\\midrule
    \bottomrule
\end{tabular}
\end{table}

Application of measurements and MMD to graphs of this scale is unlikely to properly assess generative performance on the small-scale.
To this end we apply Louvain community detection again, with a resolution of \np{1}, to both the original Cora and Facebook graphs and their sampled counterparts from \method and BTER.
Following this partitioning we calculate MMD scores on the resulting real and synthetic partitions.
QQ plots and visualisations of partitions are found in  Figure~\ref{fig:fb_synth_real_comm} for the Facebook graphs and in the supplemental material for the Cora graphs.

% \subsubsection{Stochastic Block Model}

BTER fails to replicate the essential structure of the SBM graphs, as seen in Figure \ref{fig:sbm-graphs}, but as expected does achieve close matches on clustering and degree.
It seems that \method consistently produces higher degree and clustering values than in the original SBM graphs.
Interestingly, DiGress appears to do the inverse, consistently producing lower degree and clustering values, albeit with more realistic clustering values at higher ranges than \method.
Performance degradation from using DiGress in \method, instead of by itself, is not as significant as we might have expected.

\subsubsection{Large Graphs}

% \aod{Might re-run Facebook sampling with a less enthusiastic glue model - currently about half of the edges in this graph are inter-community, which is a lot!}

% \aod{Should we structure this by dataset or by type of metric? By dataset would start with SBM (establishing against another model) then discuss performance on the larger graphs.}

% \aod{
% Points from aggregate results:
% \begin{itemize}
%     \item BTER produces quite a few un-connected components (lower node counts)
%     \item Transitivity vs clustering - transitivity is fraction of all possible triangles that ARE triangles, clustering is the average proportion of a nodes possible triangles that are triangles. Interesting that targeting clustering doesn't guarantee this graph-level metric.
% \end{itemize}
% }

We find that BTER produces a reasonable number of connected components, with using the largest component resulting in the lower number of nodes.
On whole graph measurements we find that BTER slightly outperforms \method, but not consistently on the same metrics across datasets.
The exception, where \method produces consistently more realistic distributions than BTER, is when Louvain segmentation is applied.
Here BTER produces many smaller partitions, indicating many high-modularity groupings of nodes, instead of the larger groupings produced by the original graphs and \method.
On both the Cora and Facebook graphs BTER produces more realistic clustering coefficient distributions than \method. On these large graphs \method produces nodes with lower clustering coefficients than the original graph and BTER does the opposite.

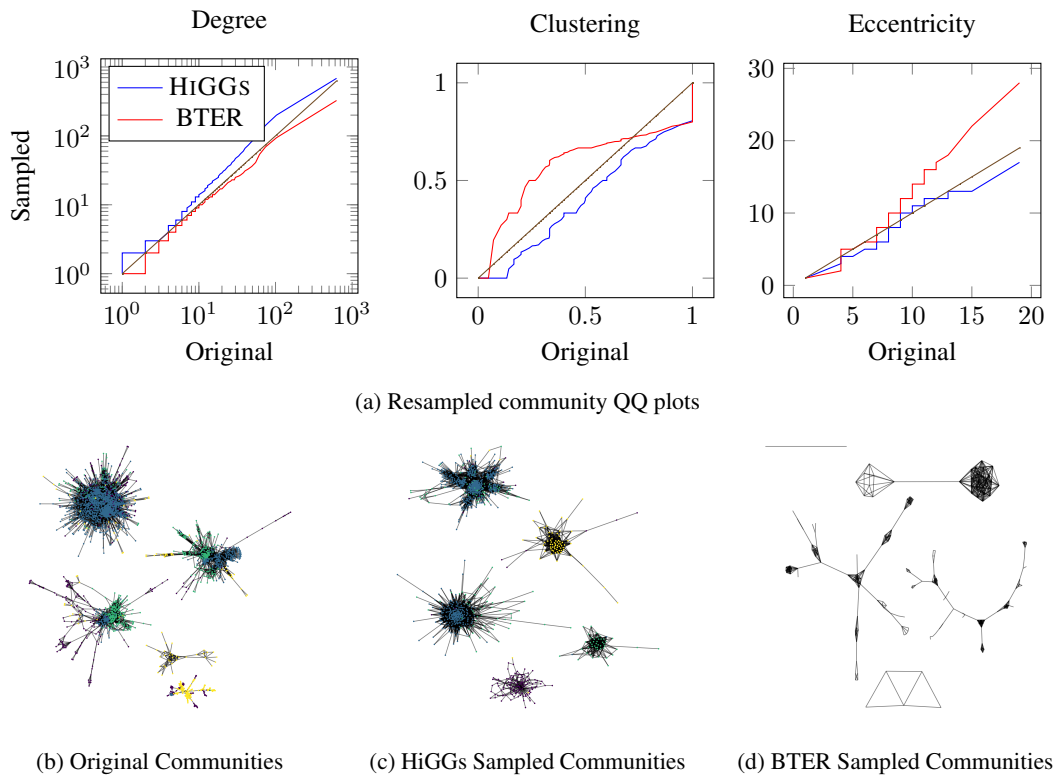
\begin{figure}[ht]
    \centering
    \begin{subfigure}[b]{\textwidth}
        \centering
            \begin{tikzpicture}
        \begin{loglogaxis}[
        title={Degree},
        height=4.7cm,
        width=5cm,
        xlabel={Original},
        ylabel={Sampled},
        legend pos=north west,
        ]
        \addplot +[mark size=0,] table [x=Degree_quantiles_real, y=Degree_quantiles_higgs, col sep=comma]
        {data/FB_QQ_Comm.csv};
        \addplot +[mark size=0,] table [x=Degree_quantiles_real, y=Degree_quantiles_bter, col sep=comma]
        {data/FB_QQ_Comm.csv};
        \addplot +[mark size=0,] table [x=Degree_quantiles_real, y=Degree_quantiles_real, col sep=comma]
        {data/FB_QQ_Comm.csv};
        \legend{\method, BTER}
        \end{loglogaxis}
        \end{tikzpicture}
    ~~
        \begin{tikzpicture}
        \begin{axis}[
        title={Clustering},
        height=4.7cm,
        width=5cm,
        xlabel={Original},
        ylabel={},
        ]
        \addplot +[mark size=0,] table [x=Clustering_quantiles_real, y=Clustering_quantiles_higgs, col sep=comma]
        {data/FB_QQ_Comm.csv};
        \addplot +[mark size=0,] table [x=Clustering_quantiles_real, y=Clustering_quantiles_bter, col sep=comma]
        {data/FB_QQ_Comm.csv};
        \addplot +[mark size=0,] table [x=Clustering_quantiles_real, y=Clustering_quantiles_real, col sep=comma]
        {data/FB_QQ_Comm.csv};
        % \legend{\method, BTER, Real}
        \end{axis}
        \end{tikzpicture}   
    ~~
            \begin{tikzpicture}
        \begin{axis}[
        title={Eccentricity},
        height=4.7cm,
        width=5cm,
        xlabel={Original},
        ylabel={},
        ]
        \addplot +[mark size=0,] table [x=Eccentricity_quantiles_real, y=Eccentricity_quantiles_higgs, col sep=comma]
        {data/FB_QQ_Comm.csv};
        \addplot +[mark size=0,] table [x=Eccentricity_quantiles_real, y=Eccentricity_quantiles_bter, col sep=comma]
        {data/FB_QQ_Comm.csv};
        \addplot +[mark size=0,] table [x=Eccentricity_quantiles_real, y=Eccentricity_quantiles_real, col sep=comma]
        {data/FB_QQ_Comm.csv};
        % \legend{\method, BTER, Real}
        \end{axis}
        \end{tikzpicture}     
        \caption{Resampled community QQ plots}
        \label{fig:fb_qq_comm}
    \end{subfigure}
    
     \centering
     \begin{subfigure}[b]{0.3\textwidth}
         \centering
         \includegraphics[width=\textwidth]{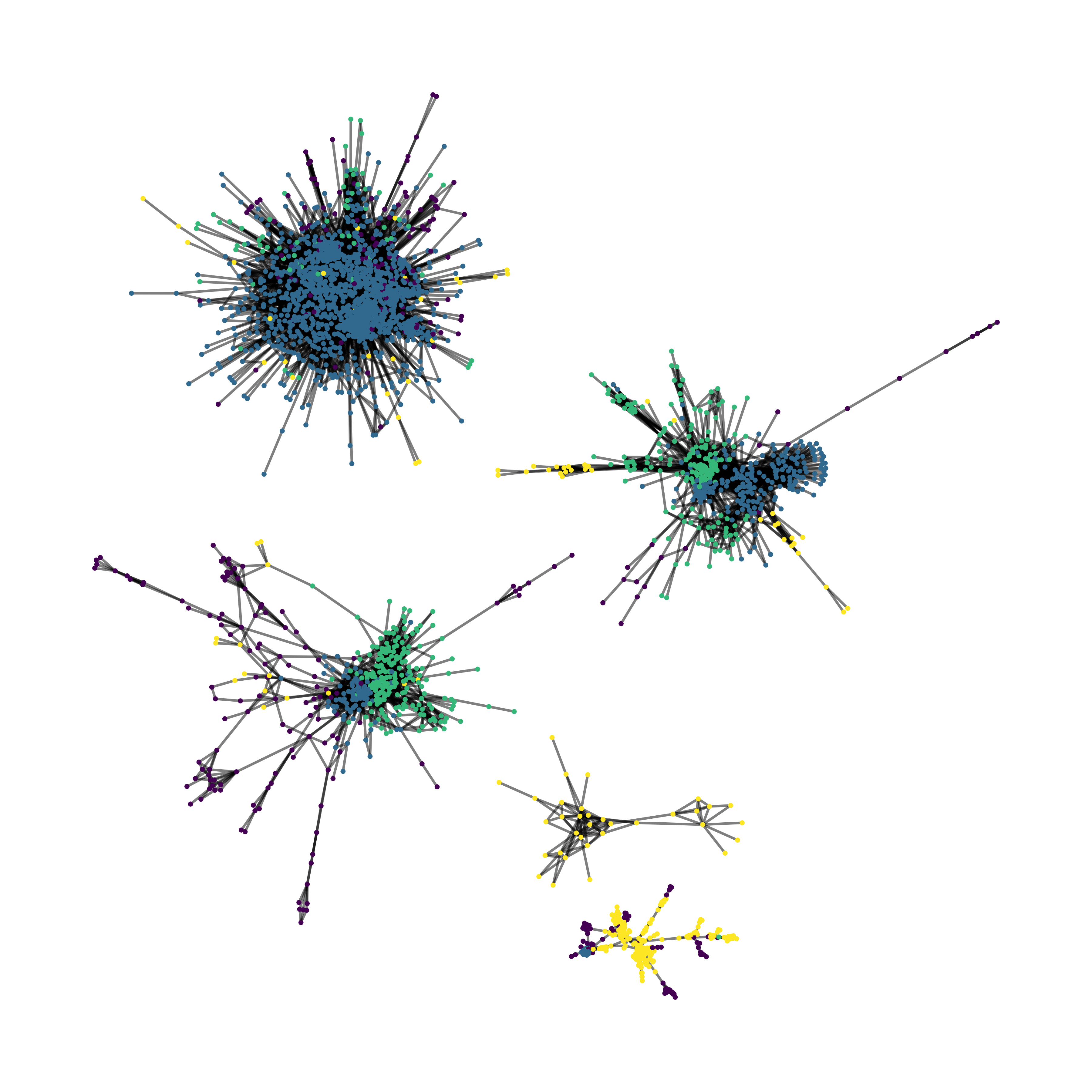}
         \caption{Original Communities}
         \label{fig:fb_real_comm}
     \end{subfigure}
     \hfill
     \begin{subfigure}[b]{0.3\textwidth}
         \centering
         \includegraphics[width=\textwidth]{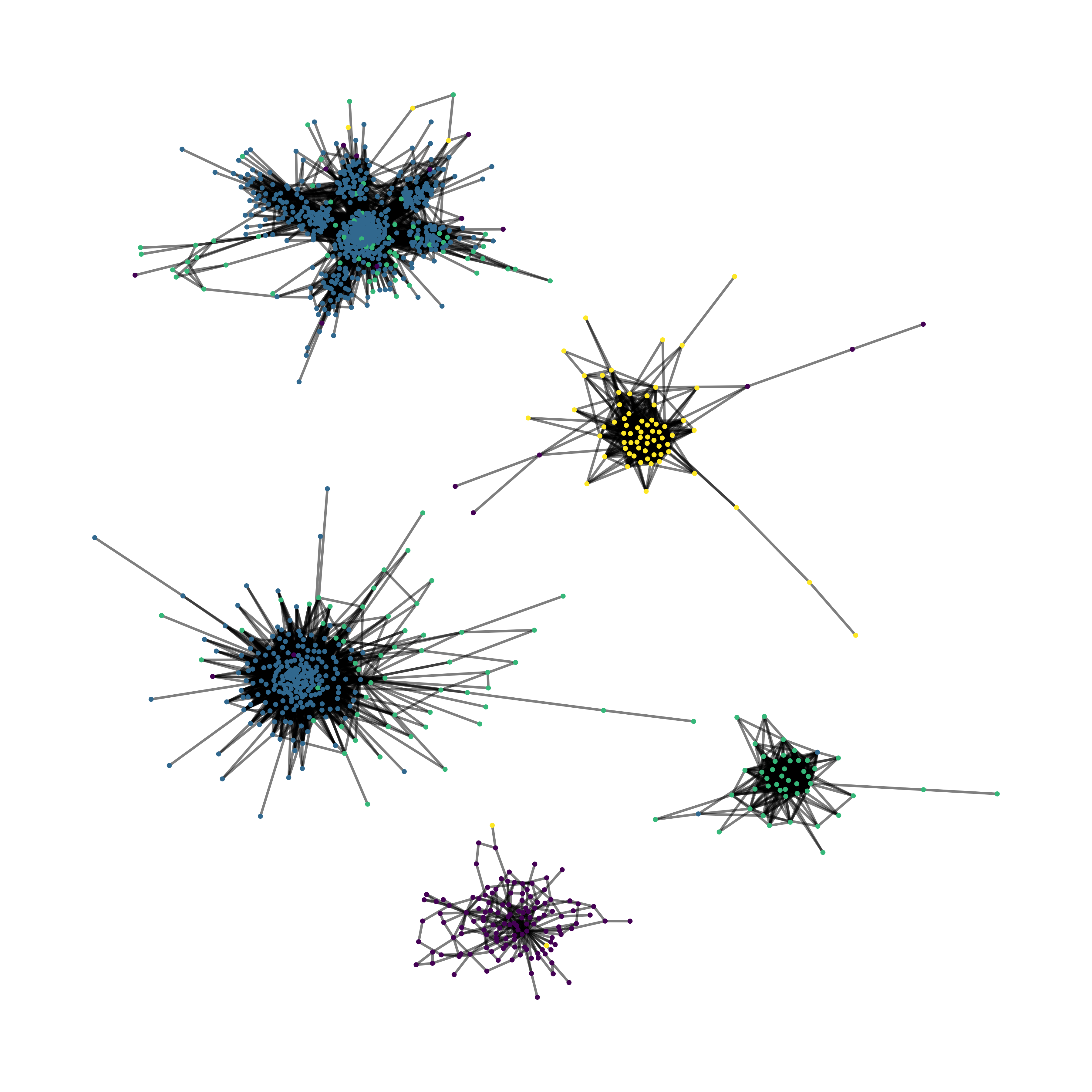}
         \caption{HiGGs Sampled Communities}
         \label{fig:fb_higgs_comm}
     \end{subfigure}
     \hfill
     \begin{subfigure}[b]{0.3\textwidth}
         \centering
         \includegraphics[width=\textwidth]{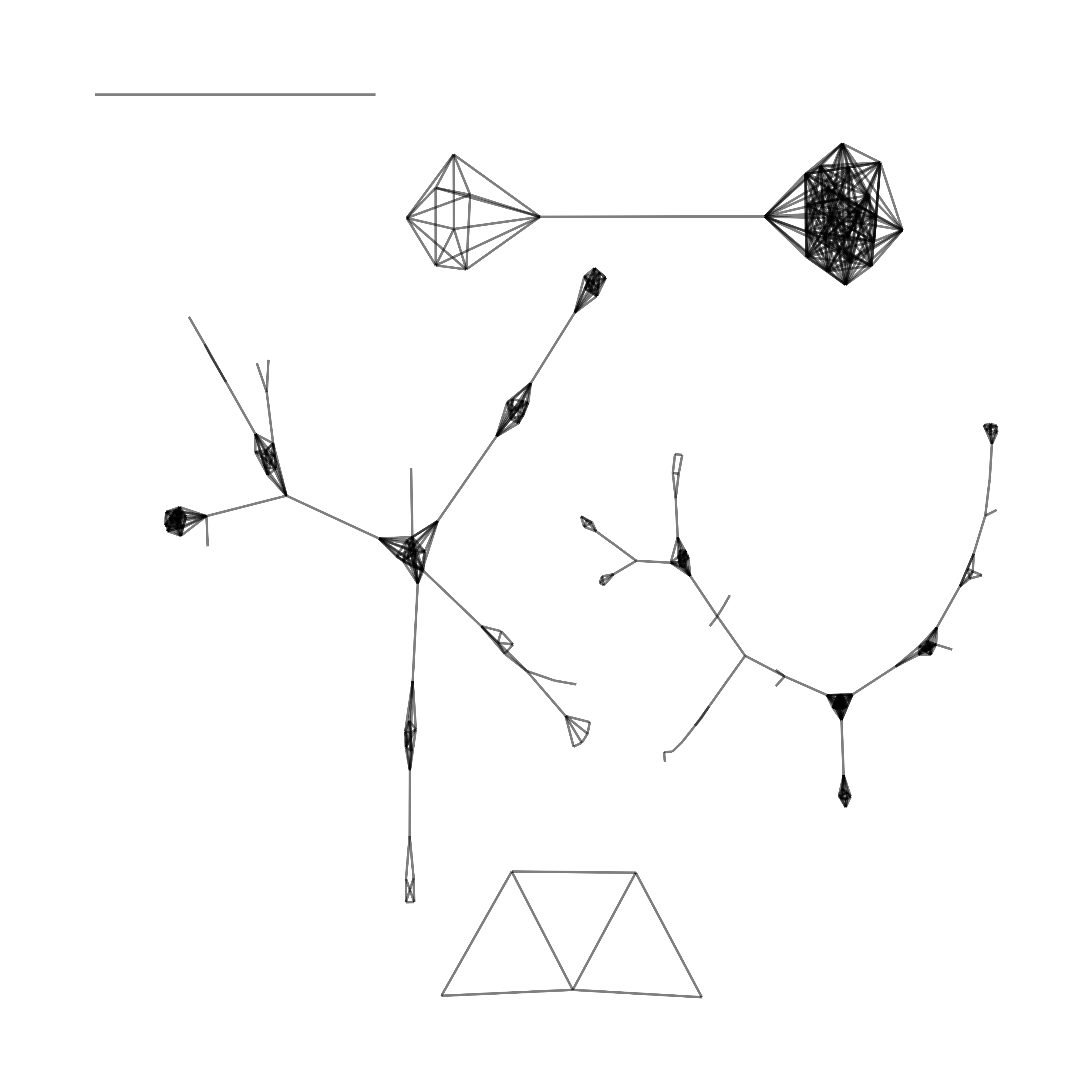}
         \caption{BTER Sampled Communities}
         \label{fig:fb_bter_comm}
     \end{subfigure}
        \caption{Communities sampled from the Facebook Page-Page graph with equivalents from \method and BTER, alongside node-level QQ plots. Layouts are through SFDP and GraphViz.}
        \label{fig:fb_synth_real_comm}
\end{figure}

Degree distributions produced by \method are much closer to the original on the Cora graph than the Facebook graph.
% , with \method out-performing BTER on the Cora graph, while BTER out-performs \method on the Facebook graph.
Indeed on the Facebook data \method over-produces high-degree nodes to the extent that the sampled graph's overall density is nearly double that of the original.
MMD scores on Louvain communities from the Facebook and Cora graphs show stronger performance on a small scale from \method than BTER.
This is particularly true on the Facebook dataset, where \method communities out-perform BTER communities on all except the distribution of community sizes 
% \footnote{Worth noting is that here the number of BTER communities is far higher which may influence MMD scores.}.
Visualisation of the communities shows clearly the prioritisation by BTER of clustering coefficients over overall structure.
% Indeed many rule-based methods like BTER, especially when targetting a specific metric like clustering, can fail to show these more complex topological features \cite{Davies2022RealisticNetworks}.
The Spectre MMD score in Table~\ref{tab:mmd_results} confirms these conclusions.

%=================================================================

% \begin{figure}[h]
%     \centering
%     \includegraphics[width=\linewidth]{images/cora_comparison.png}
%     \caption{Degree and clustering histograms, with residuals, for the CORA dataset \& its synthetic version.}
%     \label{fig:cora_deg_sample}
% \end{figure}

% \begin{figure}[h]
%     \centering
%     \includegraphics[width=\linewidth]{images/fb_comparison.png}
%     \caption{Degree and clustering histograms, with residuals, for the Facebook Page-Page dataset \& its synthetic version.}
%     \label{fig:fb_deg_sample}
% \end{figure}

% \begin{figure}[h]
%     \centering
%     \includegraphics[width=\linewidth]{images/FB.png}
%     \caption{Degree and clustering QQ plots for the Facebook Page-Page dataset \& its synthetic version.}
%     \label{fig:fb_qq}
% \end{figure}

\section{Discussion}
\label{sec:discussion}

% Here we provide a discussion of the main findings of this work, as well as our assumptions, and detail the possible risks to the validity of these findings.

% \subsection{Findings}

% This means that, as the aim of \method is explictly to produce graphs much larger those already possible with existing models, a degradation in performance at the range of graph sizes possible with other models is acceptable.

% In comparing our \method implementation to both its component models and a rule-based method, we are able to measure the extent to which \method reduces high-resolution performance, and draw comparisons to benchmarking on larger graphs.
We find that performance does drop on the SBM dataset compared to DiGress, with \method producing nodes with higher degrees and lower clustering coefficients, but this drop in quality is not massive.
At this scale of graph \method is on-par with BTER but visualisation (see supplemental material) indicates that the essential nature of these graphs has not been captured by BTER.

% On larger graphs, which (to our knowledge) no existing GNN model can extend to, benchmarking models are all rule-based.
% These seldom include node or edge attribution, especially when their focus is targeting a specific graph characteristic.
An important consideration when discussing the findings of this work is that neither benchmark model (BTER and DiGress) is necessarily a fair comparison.
Due to the in-memory costs of DiGress (and other GNN models) graphs of the scale used in our experiments simply cannot be produced.
It is arguable that rule-based models do not have the same aim as GNN models or \method.
Rule-based models generally attempt to re-construct a graph, often taking a certain distribution like degree (as in BTER) as an input parameter, and act as an exploration of mathematical modelling more than as a generative task.
This means that on quantitative metrics like MMD they often perform very strongly, but on characteristics more difficult to express through metrics, they tend to fail.

We find that on the small scale, through application of Louvain partitioning, \method has out-performed BTER.
This includes on the clustering coefficients specifically targeted by BTER, with clustering coefficients on the community scale produced by BTER consistently too high.
Visualisation of communities corroborates this, with BTER producing many triangles, instead of many-node patterns.
Conversely our \method implementation performs much better in this regard.
Given that MMD scores and aggregate metrics both rely on simple node-level measurements on graphs of this scale they may not be sufficiently expressive.
Issues with MMD, even on the few-hundred-node scale, are discussed more specifically in \citet{OBray2021EvaluationSolutions}.
We further discuss our assumptions and threats to validity in the supplemental material.

% \subsubsection{Use of Benchmark Models}

% \subsubsection{Application on Geometric Graphs}

\section{Conclusion}
\label{sec:conclusion}

In this work, we propose \method as a framework for generating large graphs. 
We implement \method using DiGress \cite{Vignac2023DiGress:Generation}, a recent graph diffusion model with the capacity to consider node attribution and conditional generation, and for inter-$h_1$ edge sampling we implement the first edge-predictive-diffusion model as a variant of DiGress.
This is, to our knowledge, the first deep-learning based method to produce graphs of tens of thousands of nodes.
Further, \method is able to do so with categorical node and edge attributes, with our generated graphs up to \np{200} times larger than those in the original DiGress paper.
Using this \method implementation, we demonstrate that on the benchmark SBM dataset performance is not significantly degraded from using DiGress by itself.
% We then evaluate our \method implementation on the Cora ($|V|=2810$) and Facebook Page-Page datasets ($|V|=22470$), which are each well beyond the current limitations of GNN models, especially with node or edge attribution.
% We benchmark against a rule-based method BTER \cite{Seshadhri2012CommunityGraphs}, which does not produce node or edge attributes.
Comparison to BTER, a rule-based method, shows that our \method implementation out-performs BTER on small-scale measurement of large sampled graphs.
However, due to BTER's use of a prescribed degree distribution, BTER out-performs our \method implementation when considering whole-graph node metric distributions.
% Qualitative analysis through visualisation of communities in the large sampled graphs suggests that \method implementation might have actually performed better than BTER.

While the implementation of \method presented here is viable for social networks, and possibly other applications, future work could focus on a few key areas. 
Firstly, whether \method is improved compared to our implementation here by using different partitioning algorithms and generative models, in particular a more efficient edge sampling model.
Secondly, an improved edge-sampling stage, that considers previous edge-samplings, might significantly improve results and allow extension to geometric or quasi-geometric graphs like molecules.
We hope that domain experts in fields such as chemistry develop their own task-specific implementations.

\aod{``We can apply this here, here and here''}


\clearpage

\appendix

\section{Appendix}

\subsection{Code Availability}

All code is available at \url{https://github.com/higgs-neurips-23/HiGGs}. This repository also contains scripts to download the trained models for each dataset. Model training and sampling was conducted on a single workstation computer equipped with an NVIDIA A6000 48gb GPU.

\subsection{\method schematic}

\begin{figure*}[ht]
    \centering
    \includegraphics[width=0.65\linewidth]{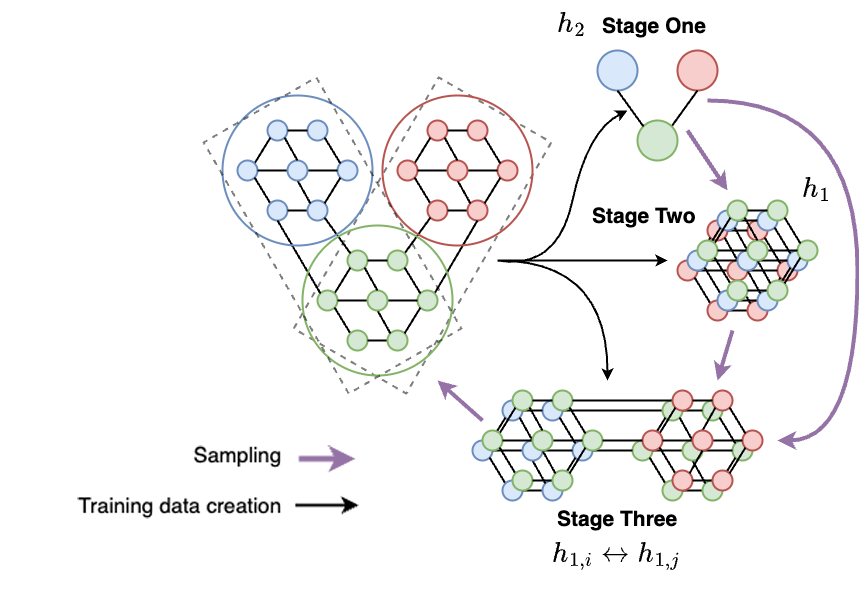}
    \caption{A basic schematic of the \method method. 
    % First an $h_2$ graph is sampled, which represents $h_1$ graph types and \textit{``are they connected''}-level inter-$h_1$ edges. Secondly each $h_1$ graph is sampled, conditioned on its type in the preceeding $h_1$ graph. Lastly, for each edge in $h_2$, the edges between the corresponding $h_1$ graphs are sampled. The result is a final large (connected) graph.
    }
    \label{fig:process_diagram_pict}
\end{figure*}

\subsection{Current Models}

\begin{table}[ht]
    \centering 
    \caption{Existing GNN graph generation methods. Order is listed as the worst case stated in the respective work and the maximum size of graphs $\lim(|V|)$ is the highest stated in the respective work, although maximum graph size was not necessarily their primary aim. $^{\dagger}$ indicates an order not explicitly stated in the respective work. \aod{Detail selection process?}
    % Cond.: Whether the model supports conditional generation.
    }
    \begin{tabular}{lcn{6}{0}cccc}\toprule
    \multirow{2}{*}{Model}   & \multirow{2}{*}{Order}            &    {\multirow{2}{*}{$\lim(|V|)$}}          & \multicolumn{3}{c}{Attribution}   &  \multirow{2}{*}{Conditioning}            \\\cmidrule{4-6}
         &           &  & Node  & Edge  & Graph     &    \\\midrule
    GraphVAE  \cite{Simonovsky2018GraphVAE:Autoencoders}     & $N^4$                  & 38            &  \cmark   &  \cmark   &  \cmark       &    \cmark \\
    DiGress         \cite{Vignac2023DiGress:Generation}             & $N^2$                  & 100           &  \cmark   &  \cmark   &  \cmark       &    \cmark \\
    GraphGDP        \cite{Huang2022GraphGDP:Generation}            & -- $^{\dagger}$                    & 399           &  \xmark    &  \xmark    &  \xmark        &    \cmark \\
    NVDiff          \cite{Chen2022NVDiff:Vectors}                  & $N$                    & 400           &  \cmark   &  \cmark   &  \xmark        &    \xmark \\
    GraphRNN        \cite{You2018GraphRNN:Models}                  & $N^2$                  & 500           &  \xmark    &  \xmark    &  \xmark        &    \xmark \\
    NetGAN          \cite{Bojchevski2018NetGAN:Walks}              & -- $^{\dagger}$                    & 2810          &  \xmark    &  \xmark    &  \xmark        &    \xmark \\
    GRAN            \cite{Liao2019EfficientNetworks}               & $N\log(N)$             & 3530          &  \xmark    &  \xmark    &  \xmark        &    \xmark \\
    \method                                                                             & See paper                   & 22000         &  \cmark   &  \cmark   &  \cmark       &    \cmark \\\bottomrule
    \end{tabular}
    \label{tab:gnn_methods}
\end{table}

\subsection{\method Complexity} 
\label{sec:order}

% \nsa{Is it Complexity divided by Order or Complexity or Order or Complexity and Order? I suggest not using "/" to indicate two words with same meaning, instead use Or or And where appropriate. }
% The order of \method varies by stage.
% Detail on the complexities of \method, independent of the models used in each stage, can be found in the supplemental material.
The size of the final sampled graph is denoted $N_G = |V|$. 
We denote the size of the $h_2$ graph $N_2 = |h_2|$.
$N_1$ is the mean size of an $h_1$ graph, $N_1= \textrm{mean}(|h_1|)= N_G / N_2$. 
The order of DiGress is $O(N^2)$, which is costly. Fortunately \method does not need to scale beyond a few hundred nodes in any given sampling process. Similarly our edge-predictive variant of DiGress scales $O(N^2)$. This means that the complexities of \method, in our implementation using Louvain \cite{DeMeo2011GeneralizedNetworks} Segmentation, DiGress, and edge-DiGress, are $O(N_G^2)$ in the segmentation stage, for both time (full sampling) and memory (per item), $O(N_2^2)$ in Stage One ($h_2$), $O(N_G^2 N_2)$ (time) and $O((N_G/N_2)^2)$ (memory) for Stage Two ($h_1$), and $O(N_G^2)$  and $O((N_G/N_2)^2)$ for Stage Three ($h_{1,i} \leftrightarrow h_{1, j}$).
% This means that \method, in our implementation using DiGress and edge-DiGress, has complexities shown in Table~\ref{tab:complexities}.
The maximum memory requirement of edge-sampling is likely the limiting factor, because the largest graphs passed to this model are double that of the largest from $h_1$. 

The order of \method varies by stage (see above). 
The size of the final sampled graph is denoted $N_G = |V|$. 
We denote the size of the $h_2$ graph $N_2 = |h_2|$, ie the cardinality of its node-set.
Similarly $N_1$ is the mean size of an $h_1$ graph, $N_1= \textrm{mean}(|h_1|)$. 
The number of edges in $h_2$ is $E_2 = |E_{h_2}|$.
% \aod{Definitely needs cleaner notation}. 
The order of each stage is necessarily very dependent on the order of the generative model being used. 
% Some models, for example GRAN from \cite{Liao2019EfficientNetworks}, are $O(N\log(N)$. 
% Most models, in particular those that consider edge and node classes, are at least $O(N^2)$. 
We denote the order of an arbitrary model as $O(N^{x, y, z})$ with $x, y, z$ an unknown (assumed polynomial) complexity for the models in Stage One, Two and Three respectively.
Training complexities for each stage simply inherit from the model, ie each stage has order $O(N_{\textrm{stage}}^i)$. 

The order of $h_2$ sampling simply inherits the order of the generative model, ie $O(h_2) = O(N_2^x)$.
% \subsubsection{$h_1$ Sampling}
An $h_1$ graph is sampled for each node in the proceeding $h_2$ graph, 
% and the time complexity of sampling for that $h_1$ graph inherits from the model used, 
so Stage 2 has sampling time complexity $O(h_1) = O(N_2 N_1^y)$. We can expect an average community size dependent on the size of the final sampled graph, so $N_1 = N_G/N_2$, and $O(h_1)=O(N_G^y N_2^{y-1})$.

% \subsubsection{$h_{1,i} \leftrightarrow h_{1, j}$ Sampling}

% We make the assumption here that complexity is again inherited from the same model as in the previous stages - however, edge prediction models with lower order are more common, so this is not necessarily a given.
The upper limit on size of graphs during the edge-sampling stage is double those from $h_1$.
The number of samples is determined by the number of edges in $h_2$, $E_2 = \rho (N_2^2/2 - N)$, where $\rho$ is the density of edges in $h_2$.

This leads to an order of $O(h_{1,i} \leftrightarrow h_{1, j}) = O(N_2^2 N_1^z)$. 
Note here however that memory capacity of the GPU used is the likely limit, as we might expect $N_1^x << (2N_1)^z$. 
Using the above $N_1 = N_G/N_2$, this becomes $O(h_{1,i} \leftrightarrow h_{1, j}) = O(N_G^z N_2^{2-z})$, with the actual number of edge-prediction operations scaled linearly by $h_2$ edge density $\rho_2$.
Crucially, and unlike block-generative or motif-based models, stages two and three consist of many independent jobs, making them trivially paralellisable.

\subsection{Extra Features}

\subsubsection{Diameter}

As we are extending the number of nodes from $|V|\sim 100$ to $|V| \sim 600$ in some cases, the planarity/diameter of the graph becomes less easily expressed by DiGress. With the aim of alleviating some of these issues we implement a simple approximation for graph diameter, with the same principles as MultiBFS from \citet{Boitmanis2006FastGraph}.

In brief summary, for a given message passing kernel size $n$, a simple convolution without weights is given by Equation~\ref{eqn:Pn}.

\begin{multicols}{3}

\begin{equation}
    P_n = \sum_{i=0}^n A^i
\label{eqn:Pn}
\end{equation}

\columnbreak

\begin{equation}
    x_{i,j} = 
    \begin{cases}
    1  & \text{if } j = i\\
    0 & \text{otherwise}
    \end{cases}
\label{eqn:xn}
\end{equation}

\columnbreak

\begin{equation}
    x'_n = P_n x
\label{eqn:xdash}
\end{equation}

\end{multicols}

Selecting a random node $u$ in a graph, we construct a bit-vector $x$ with zeroes everywhere except the index of that node, as in Equation~\ref{eqn:xn} for initial node index $i$. Applying the convolution gives an updated per-node feature vector given by Equation~\ref{eqn:xdash}. Crucially this vector is zero for nodes $v$ where a message has not reached - which means that the shortest path between the $u$ and $v$ is longer than $n$, i.e., $SP(u,v) > n$. 

% By constructing several convolutions of varying kernel size (we use $n = 2, 4, 6, 8$) and repeating the application of these convolutions for multiple random initial nodes (here $5$) we can calculate the proportion $f$ of node pairs in the graph with $SP(u,v) > n$ for each $n$. $f$ should be a reasonable estimate for the average number of nodes reachable in $<n$ path length, so should serve as a feature with which the model can learn to optimise for.

By iteratively applying a $P_n$ message passing convolution to the bit-vector $x$, and tracking at what $n$ all nodes have received at least one message, we can derive the eccentricity of an individual node. We sample the eccentricities of five nodes per graph per batch then take the maximum of these as an estimate of diameter.

The advantage of this approach, as opposed to a direct diameter calculation using BFS or similar, is that it can be computed entirely through matrix operations, and on a whole batch simultaneously. This means that no movement from or to the GPU is required when it is computed during training.

% \aod{Alternative: Recursively apply until furthest node is reached. Note sure which will be quicker! Recursive application would give much better estimate of network diameter.}

\subsubsection{Clustering}

% \aod{Ended up working out how to just calculate clustering coefficient from an adjacency matrix - nice and fast and gives good indication of local clustering (although the maths needs a look-over).}

For a graph with adjacency matrix $A$, which is a bit-matrix showing the existence of any edges, we can count the number of $N$-node cycles for each node $i$, denoted $|\theta_{n,i}|$, using the trace of the matrix's product with itself, i.e., for all nodes in a graph $|\theta_n| \in \mathbb{Z}^{|V|}$ is Equation~\ref{eqn:n_cycles}.

\begin{multicols}{2}

\begin{equation}
    |\theta_{n}| = \textrm{Tr}(A^n)
    \label{eqn:n_cycles}
\end{equation}

\columnbreak

\begin{equation}
    C = \frac{|\theta_3|}{|\theta_2|} = \frac{\textrm{Tr}(A^3)}{\textrm{Tr}(A^2)(\textrm{Tr}(A^2) - 1)}
    \label{eqn:cluster}
\end{equation}

\end{multicols}

The local clustering efficient of a node is the probability that two of its neighbours are also each others neighbours.
This can be formulated as the number of possible ($3$-cycles) that are complete given the node's $2$-cycles.
$2$-cycles are in effect the degree of the node in question.
This allows gpu-based calculation of local per-node clustering $C \in \mathbb{R}^{|V|}$, from the adjacency matrix, using Equation~\ref{eqn:cluster}.
Note here that for clustering we do not include $2$-cycles in which the root note appears twice (eg $v_1 \rightarrow v_2 \rightarrow v_1$ is excluded).
We include as extra features both node-level local clustering as a node feature and the mean clustering coefficient as a graph-level feature.

\subsection{edge-DiGress}

This is constructed from an array $K = [\textrm{True, True, ..., False, False}]$ with True representing nodes in one community and False the other\footnote{This requires that nodes are not re-ordered between dataset construction and passing to eDiGress}. From this array $K$ we construct a matrix $\hat{K}$ with each row or column a repeat of $K$, ie $\hat{K}_{[:,i]} = K$ with $\hat{K} \in \mathbb{B}^{N \times N}$ \footnote{$\mathbb{B}$ signifying boolean data and $N$ the size of the adjacency matrix}. From here a node-node $h_1$ match matrix $\hat{M}$ is:

\begin{equation}
    \hat{M}_{i,j} = 
\begin{cases}
    \text{True}  & \text{if } \hat{K}_{i,j} = \hat{K}^{\top}_{i,j}\\
    \text{False} & \text{otherwise}
\end{cases}
\end{equation}

which can simply be applied as a mask to exclude intra-$h_1$ edges from noise. During sampling the noise limit is sampled and applied to any possible inter-$h_1$ edges. Node classes are kept constant at each stage of noise application and sampling, so noise is applied only on edges, and sampling after input of two $h_1$ graphs returns those same graphs with new inter-$h_1$ edges. 

\subsection{Metrics}

MMD requires the measurement of the distribution of a given quantity, with MMD then calculated between these distributions, with $\textrm{MMD}=0$ for identical distributions.
The quantities we measure and use for MMD in this work are detailed below.
 
%  \aod{Double check kernel detail}

% \aod{Give more description of each - how computed mainly, merge subsubs below}
\begin{description}
    \item[Nodes] The number of nodes in each graph, $|V|$.
    
    \item[Degree] The degree of each node, denoted $d(v)$ for a node $v$.
    
    \item[Clustering] The fraction of possible triangles through a node that exist i.e. the extent to which a node and its neighbours are a complete graph. Clustering $c_v$ for a node $v$ is calculated as 
    \begin{equation}
        c_v = \frac{2T(v)}{d(v)(d(v)-1)}
    \end{equation}
    where $T(v)$ is the number of triangles through $v$, and $d(v)$ is the degree of $v$. Dense clusters in graphs have high clustering coefficients, due to their high inter-connectedness, whereas more sparse regions have low clustering coefficients.

    \item[Eccentricities] All node-node shortest path lengths in a graph. The shortest path between two nodes is the path that which contains the fewest nodes. % (REPLACE WITH DIAMETER?)
    
    \item[Spectre] %Eigenvalues of the Laplacian of a graph. 
    For an un-directed graph $G$, we can construct a diagonal matrix of node degrees $D$. From this we can calculate the Laplacian of the graph, $L = D - A$, where $A$ is the graph's adjacency matrix. The normalised Laplacian is then:
    
    \begin{equation}
        N = D^{-1/2}LD^{-1/2}
    \end{equation}
    
    % \begin{equation}
    %     L = D - A
    % \end{equation}

    Taking the eigenvalues of this normalised Laplacian allows us to treat the graph as defined by a spectrum. Spectral analysis takes a view of global graph properties, unlike degree or clustering, which are local statistics.
\end{description}

\subsection{Results}

\begin{table}[!htb]
\centering
\caption{Details of aggregate results for our experiments. For the SBM dataset results are the mean value across all graphs generated and in the test-set. }
\label{tab:aggregate_results}

\centering
    % \begin{tabular}{c||c|c|c|c}
    \begin{tabular}{l n{5}{0} n{6}{0} n{2}{2} n{3}{0} n{3}{0} n{1}{3} n{1}{6} n{1}{3}}
    \toprule
    &   {Nodes}         &   {Edges}           &   {Diameter}  & {$N_C$} & {$|C|_{med}$}   & {Clust.}   &   {Density}    &    {Trans.} \\\midrule
    \textbf{SBM}       &    107          &     522     &   5.46        &    {--}     &    {--}           & 0.276           &   0.099       &    0.276   \\
    DiGress   &    101          &     425     &   5.90        &    {--}     &     {--}          & 0.231           &   0.0982      &    0.236  \\
    \method   &    122          &     679     &   6.08        &    {--}     &     {--}          & 0.349           &   0.0954      &    0.344  \\
    BTER      &    107          &     462     &   5.90        &    {--}     &     {--}          & 0.140           &   0.0873      &    0.121  \\\midrule
    % ====================================================================================================================
    \textbf{Cora}      &   2810          &   7981      &   17          &     17  & 110           & 0.277           &   0.00202     &    0.114   \\
    \method   &   2951          &   7981      &   18          &     29  & 94            & 0.163           &   0.00183     &    0.130  \\
    BTER      &   2019          &   5143      &   16          &     51  & 30            & 0.307           &   0.00252     &    0.271  \\\midrule
    % ====================================================================================================================
    \textbf{Facebook}  &   22470         &   171002    &   15          &     60  & 167           & 0.360           &   0.000677   &    0.232   \\
    \method   &   21643         &   283552    &   20          &     54  & 213           & 0.243           &   0.00121    &    0.130   \\
    BTER      &   19070         &   131065    &   17          &     221 &  78           & 0.397           &   0.000721   &    0.274   \\\midrule
    % Iceland   &   27980         &   33190     &   246         & 0.0472        &   0.0000847  &    0.0732  \\
    % Sampled   &   24651         &   29847     &   76          & 0.0125        &   0.0000982  &    0.0171  \\\midrule

\end{tabular}
\end{table}

\begin{figure}[ht]
    \centering
    \begin{subfigure}[b]{\textwidth}
        \centering
        \centering
    \begin{tikzpicture}
    \begin{loglogaxis}[
    title={Degree},
    height=4.7cm,
    width=5cm,
    xlabel={Original},
    ylabel={Sampled},
    legend pos=north west,
    ]
    \addplot +[mark size=0,] table [x=Degree_quantiles_real, y=Degree_quantiles_higgs, col sep=comma]
    {data/FB_QQ.csv};
    \addplot +[mark size=0,] table [x=Degree_quantiles_real, y=Degree_quantiles_bter, col sep=comma]
    {data/FB_QQ.csv};
    \addplot +[mark size=0,] table [x=Degree_quantiles_real, y=Degree_quantiles_real, col sep=comma]
    {data/FB_QQ.csv};
    \legend{\method, BTER}
    \end{loglogaxis}
    \end{tikzpicture}
~~
    \begin{tikzpicture}
    \begin{axis}[
    title={Clustering},
    height=4.7cm,
    width=5cm,
    xlabel={Original},
    ylabel={},
    ]
    \addplot +[mark size=0,] table [x=Clustering_quantiles_real, y=Clustering_quantiles_higgs, col sep=comma]
    {data/FB_QQ.csv};
    \addplot +[mark size=0,] table [x=Clustering_quantiles_real, y=Clustering_quantiles_bter, col sep=comma]
    {data/FB_QQ.csv};
    \addplot +[mark size=0,] table [x=Clustering_quantiles_real, y=Clustering_quantiles_real, col sep=comma]
    {data/FB_QQ.csv};
    % \legend{\method, BTER, Real}
    \end{axis}
    \end{tikzpicture}   
~~
        \begin{tikzpicture}
    \begin{axis}[
    title={Eccentricity},
    height=4.7cm,
    width=5cm,
    xlabel={Original},
    ylabel={},
    ]
    \addplot +[mark size=0,] table [x=Eccentricity_quantiles_real, y=Eccentricity_quantiles_higgs, col sep=comma]
    {data/FB_QQ.csv};
    \addplot +[mark size=0,] table [x=Eccentricity_quantiles_real, y=Eccentricity_quantiles_bter, col sep=comma]
    {data/FB_QQ.csv};
    \addplot +[mark size=0,] table [x=Eccentricity_quantiles_real, y=Eccentricity_quantiles_real, col sep=comma]
    {data/FB_QQ.csv};
    % \legend{\method, BTER, Real}
    \end{axis}
    \end{tikzpicture}   
        \caption{Whole-graph QQ plots}
        \label{fig:fb_qq}
    \end{subfigure}
     \centering
     \begin{subfigure}[b]{0.3\textwidth}
         \centering
         \includegraphics[width=\textwidth]{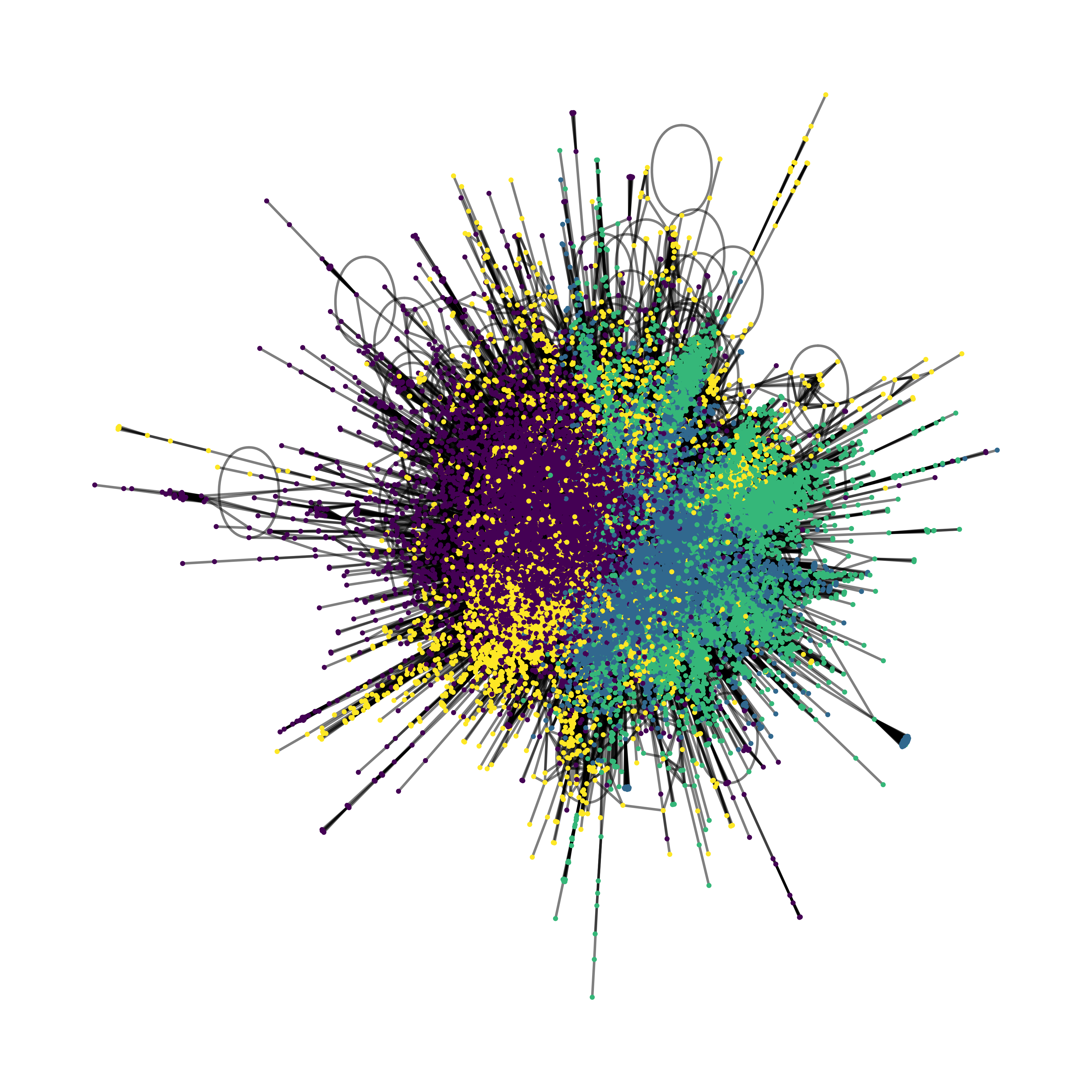}
         \caption{Original}
         \label{fig:fb_real}
     \end{subfigure}
     \hfill
     \begin{subfigure}[b]{0.3\textwidth}
         \centering
         \includegraphics[width=\textwidth]{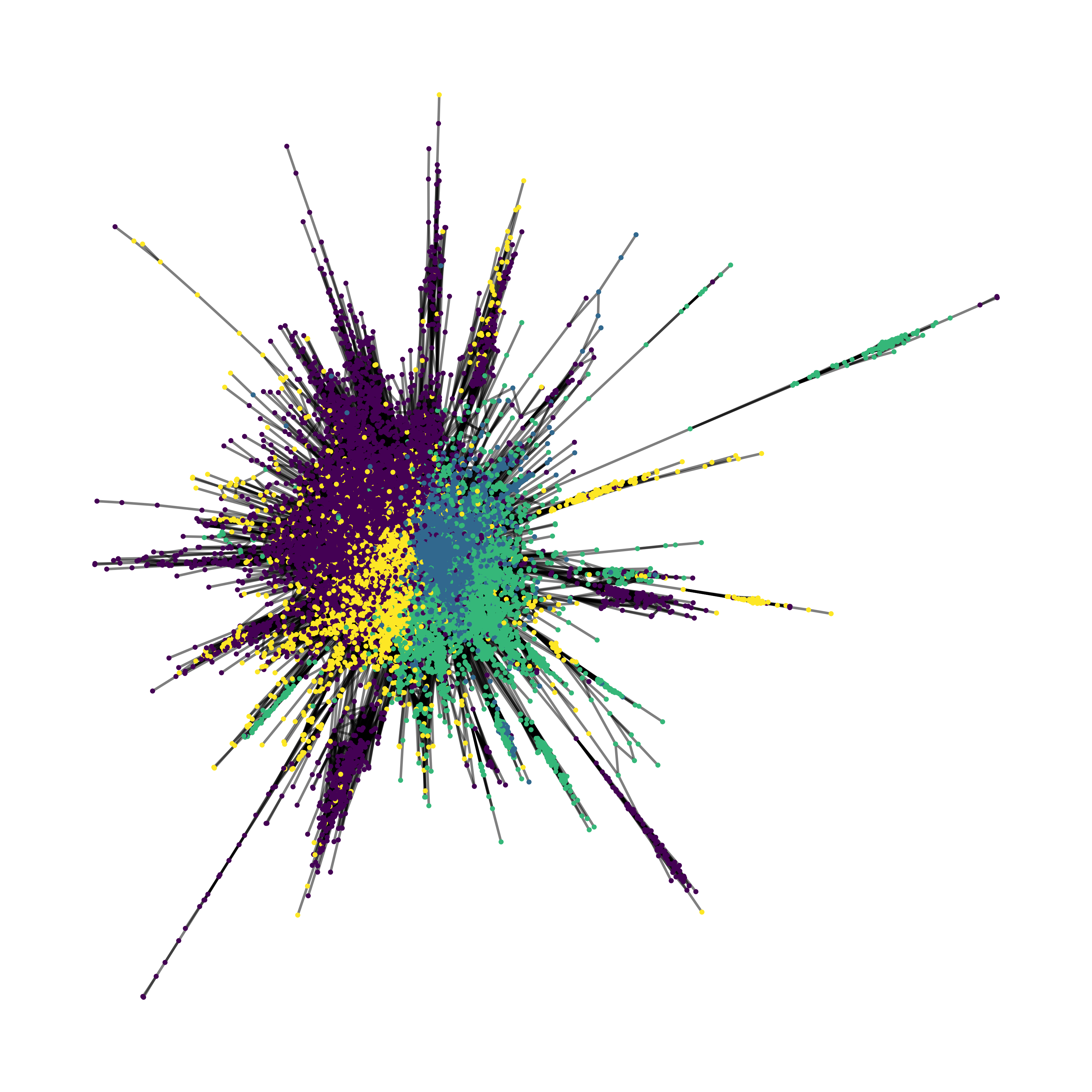}
         \caption{HiGGs Sampled}
         \label{fig:fb_higgs}
     \end{subfigure}
     \hfill
     \begin{subfigure}[b]{0.3\textwidth}
         \centering
         \includegraphics[width=\textwidth]{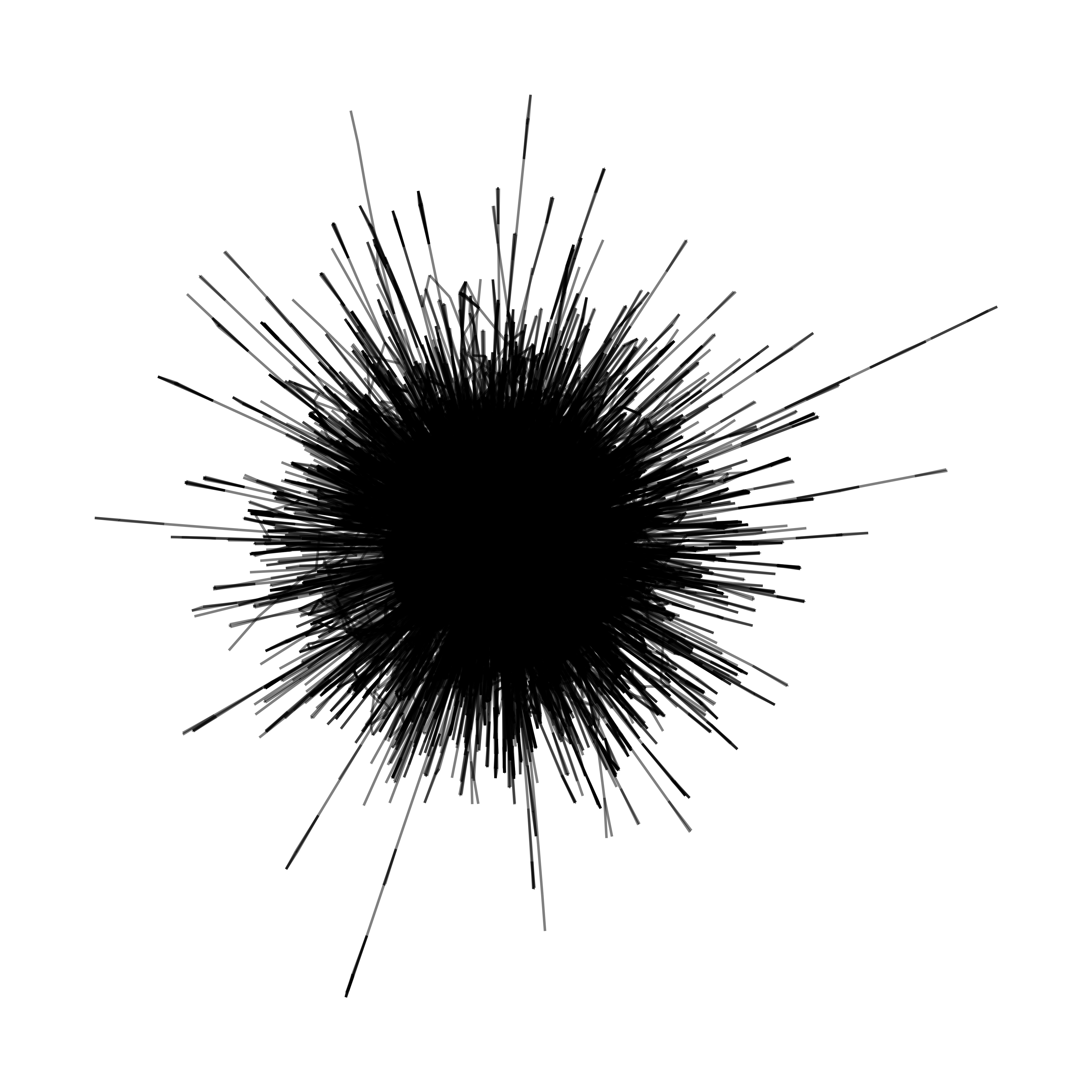}
         \caption{BTER Sampled}
         \label{fig:fb_bter}
     \end{subfigure}
        \caption{The Facebook Page-Page graph with sampled counterparts from \method and BTER, with node-level QQ plots. Layouts are through SFDP and GraphViz.}
        \label{fig:fb_synth_real}
\end{figure}

\begin{figure}[htb]
    \centering
    \begin{subfigure}[b]{\textwidth}
        \centering
        \includegraphics[width=\textwidth]{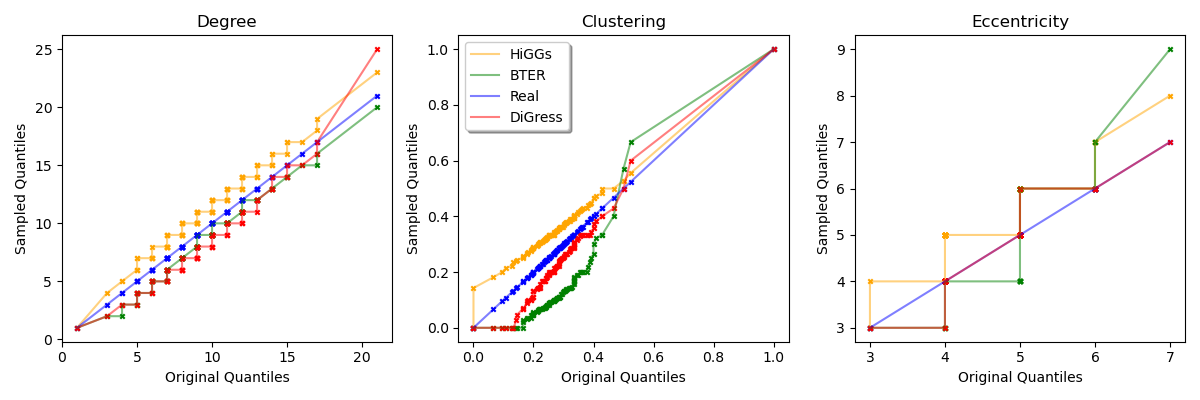}
        \label{fig:sbm_qq}
    \end{subfigure}
    \begin{subfigure}[b]{0.49\textwidth}
        \centering
        \includegraphics[width=\textwidth]{images/Real_SBM.png}
        \caption{Original}
        \label{fig:sbm_real}
    \end{subfigure}
    \hfill
     \begin{subfigure}[b]{0.49\textwidth}
         \centering
         \includegraphics[width=\textwidth]{images/Sampled_SBM.png}
         \caption{\method}
         \label{fig:sbm_higgs}
     \end{subfigure}
     \hfill
     \begin{subfigure}[b]{0.49\textwidth}
         \centering
         \includegraphics[width=\textwidth]{images/BTER_SBM.png}
         \caption{BTER}
         \label{fig:sbm_bter}
     \end{subfigure}
     \hfill
     \begin{subfigure}[b]{0.49\textwidth}
         \centering
         \includegraphics[width=\textwidth]{images/DGD_SBM.png}
         \caption{DiGress}
         \label{fig:sbm_dgd}
     \end{subfigure}
        \caption{SBM graphs, with counterparts produced by BTER, DiGress and \method.}
        \label{fig:sbm_synth_real}
\end{figure}

\begin{figure}[ht]
    \centering
    \begin{subfigure}[b]{\textwidth}
        \centering
        \includegraphics[width=\textwidth]{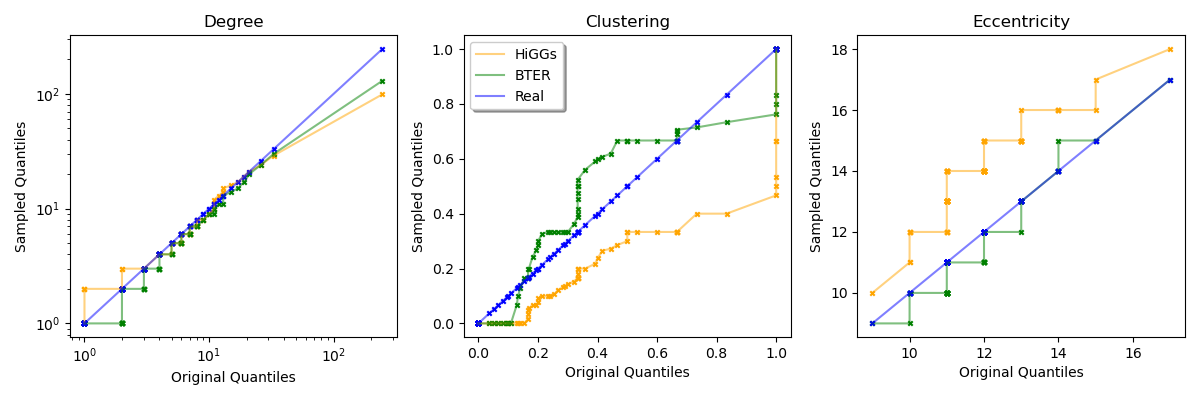}
        \caption{Whole-graph QQ plots}
        \label{fig:cora_qq}
    \end{subfigure}
     \centering
     \begin{subfigure}[b]{0.48\textwidth}
         \centering
         \includegraphics[width=\textwidth]{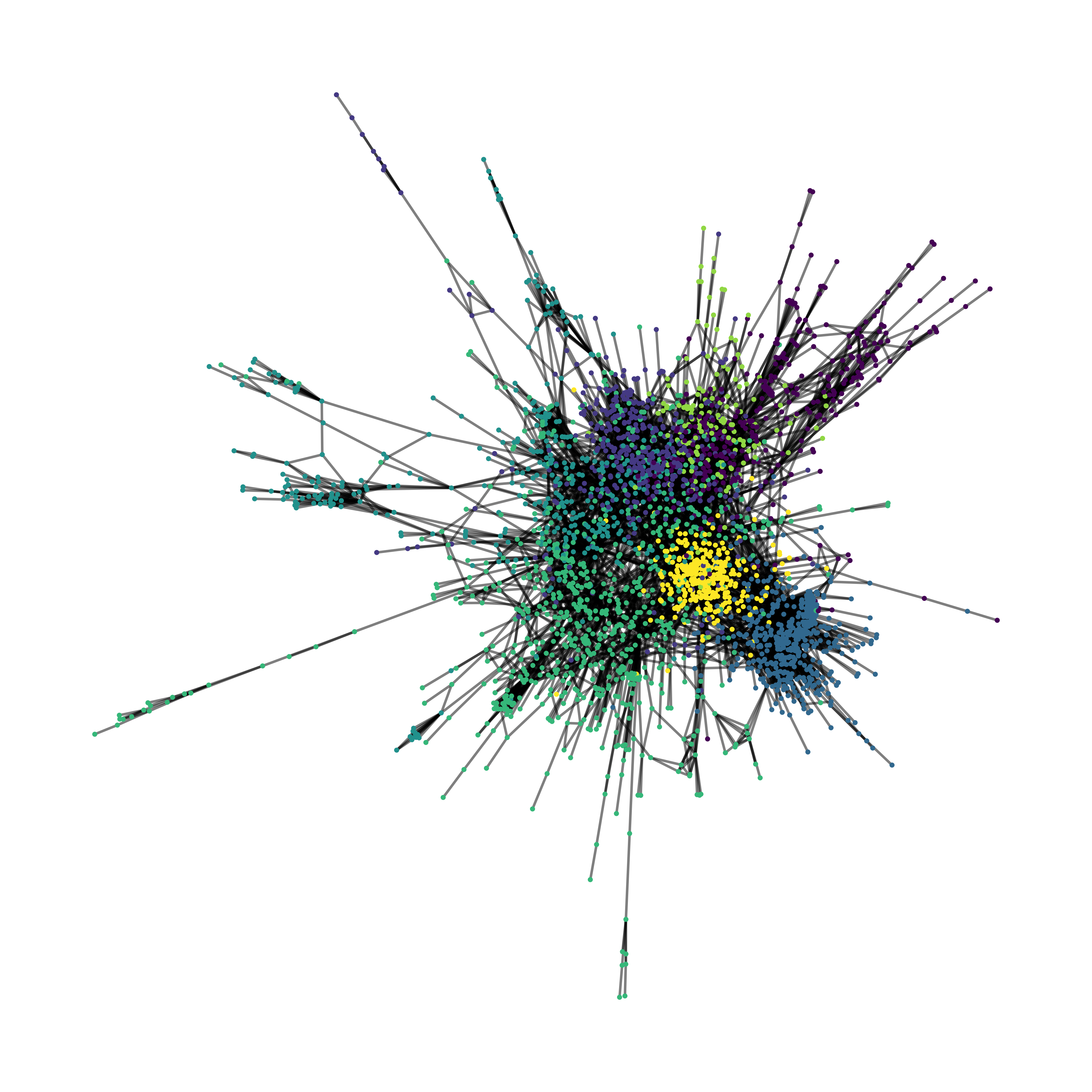}
         \caption{Original}
         \label{fig:cora_real}
     \end{subfigure}
     \hfill
     \begin{subfigure}[b]{0.48\textwidth}
         \centering
         \includegraphics[width=\textwidth]{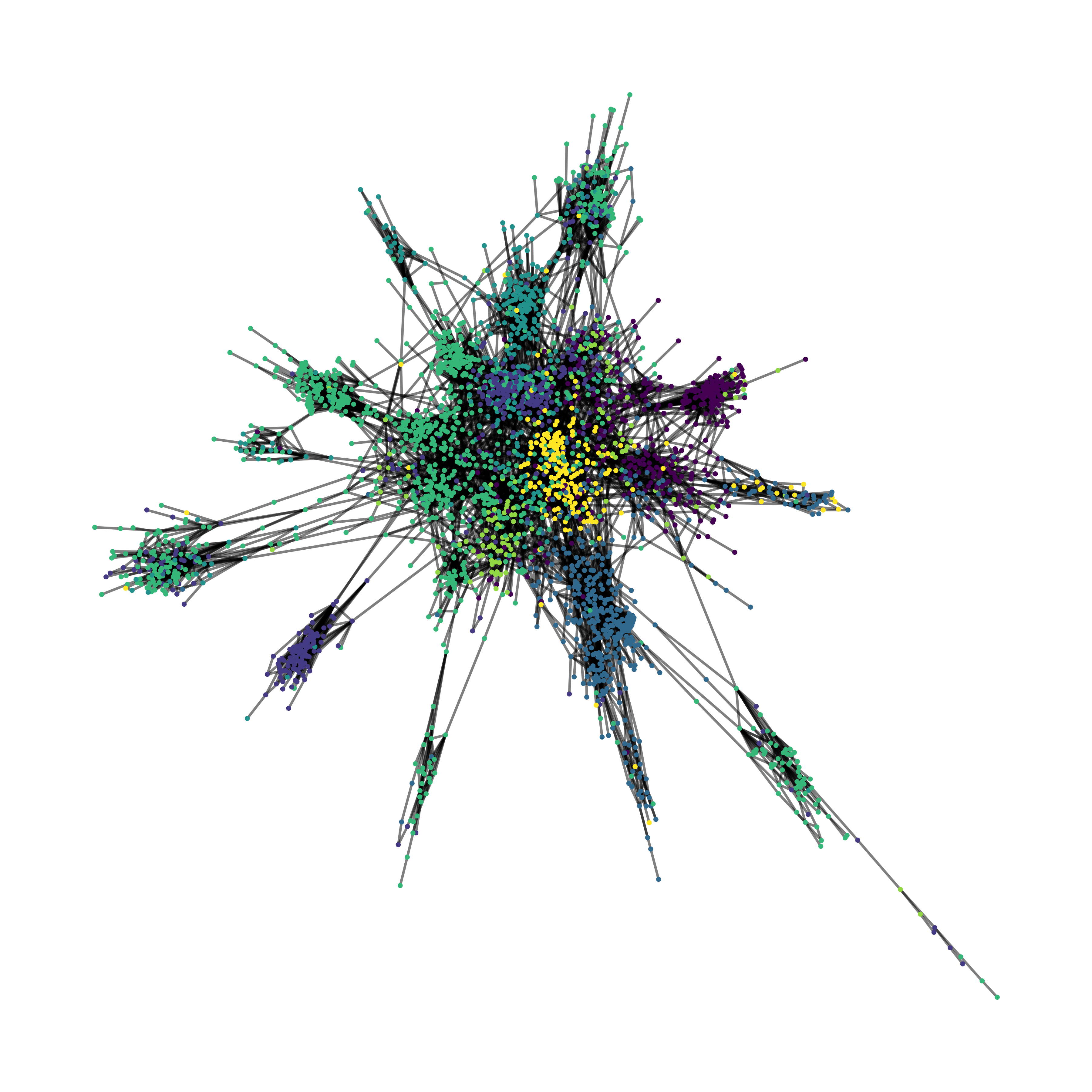}
         \caption{HiGGs Sampled}
         \label{fig:cora_higgs}
     \end{subfigure}
     % \hfill
     \centering
     \begin{subfigure}[b]{0.48\textwidth}
         \centering
         \includegraphics[width=\textwidth]{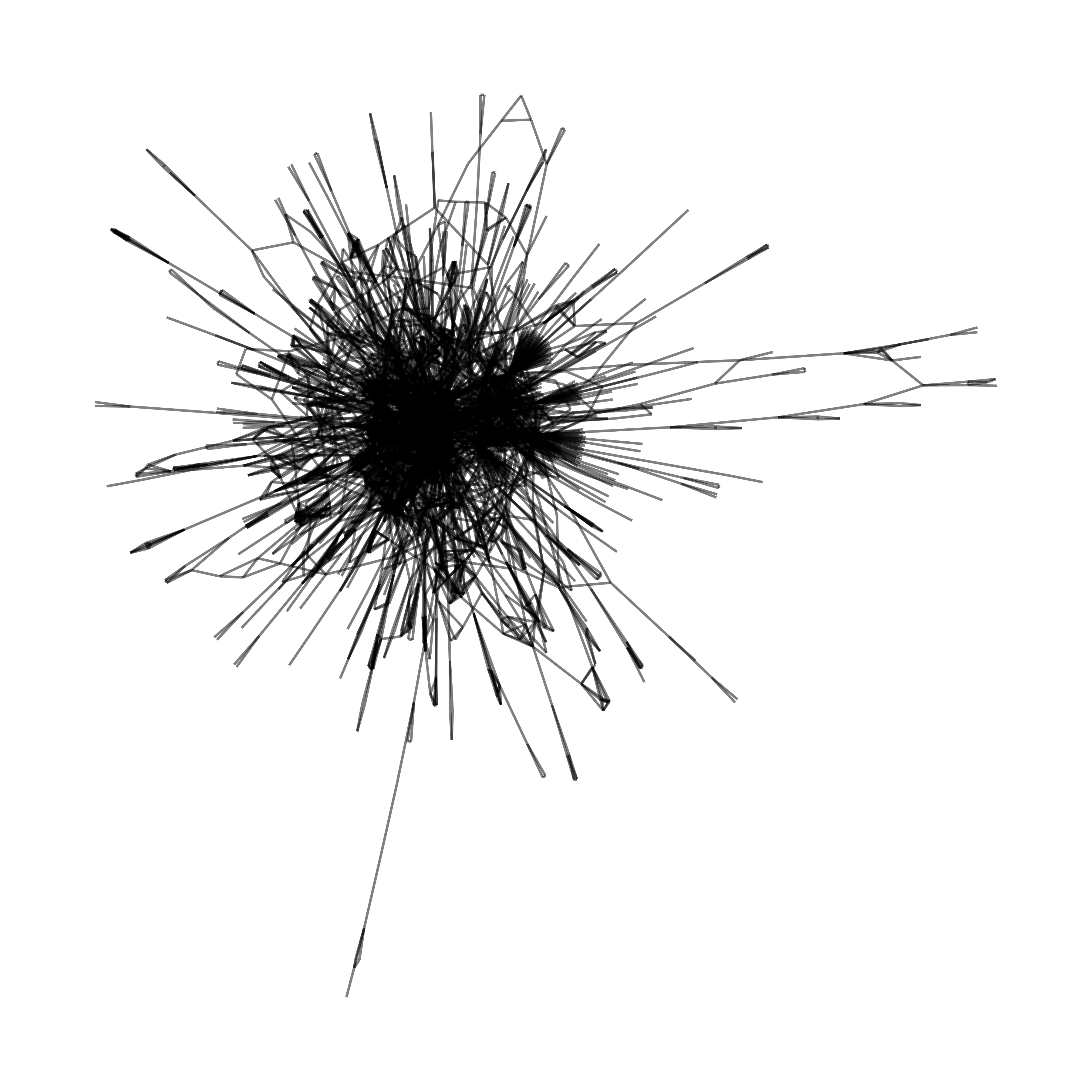}
         \caption{BTER Sampled}
         \label{fig:cora_bter}
     \end{subfigure}
        \caption{The CORA graph with sampled counterparts from \method and BTER, with node-level QQ plots. Layouts are through SFDP and GraphViz.}
        \label{fig:cora_synth_real}
\end{figure}

\begin{figure}[ht]
    \centering
    \begin{subfigure}[b]{\textwidth}
        \centering
        \includegraphics[width=\textwidth]{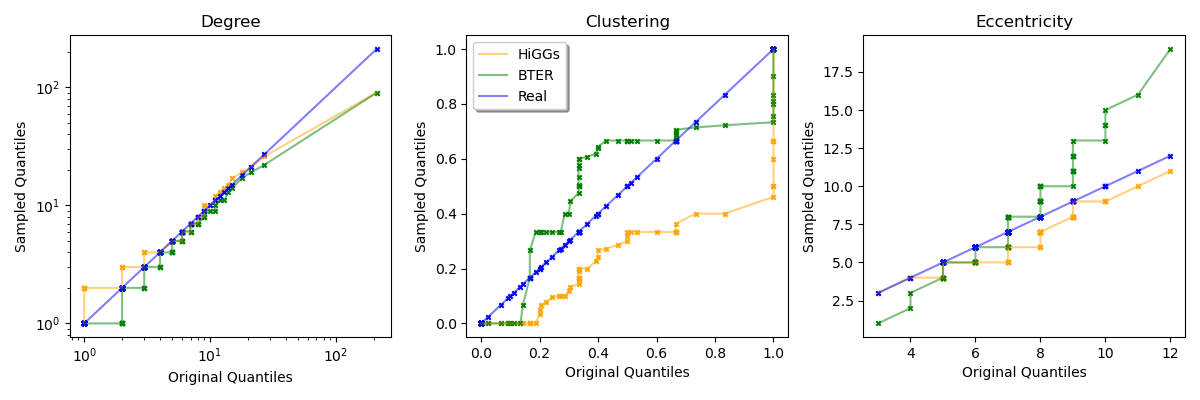}
        \caption{Resampled community QQ plots}
        \label{fig:cora_qq_comm}
    \end{subfigure}
     \centering
     \begin{subfigure}[b]{0.48\textwidth}
         \centering
         \includegraphics[width=\textwidth]{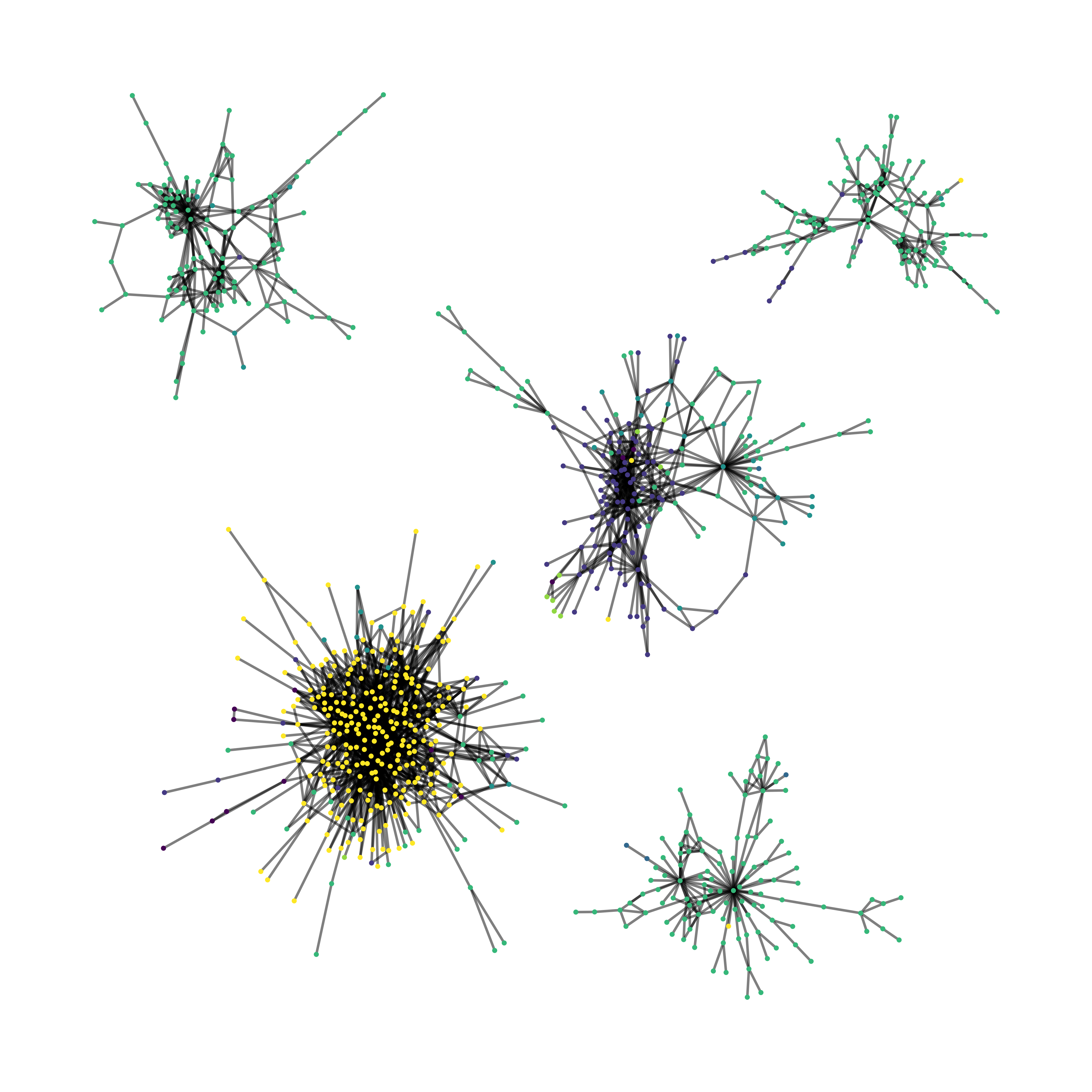}
         \caption{Original}
         \label{fig:cora_real_comm}
     \end{subfigure}
     \hfill
     \begin{subfigure}[b]{0.48\textwidth}
         \centering
         \includegraphics[width=\textwidth]{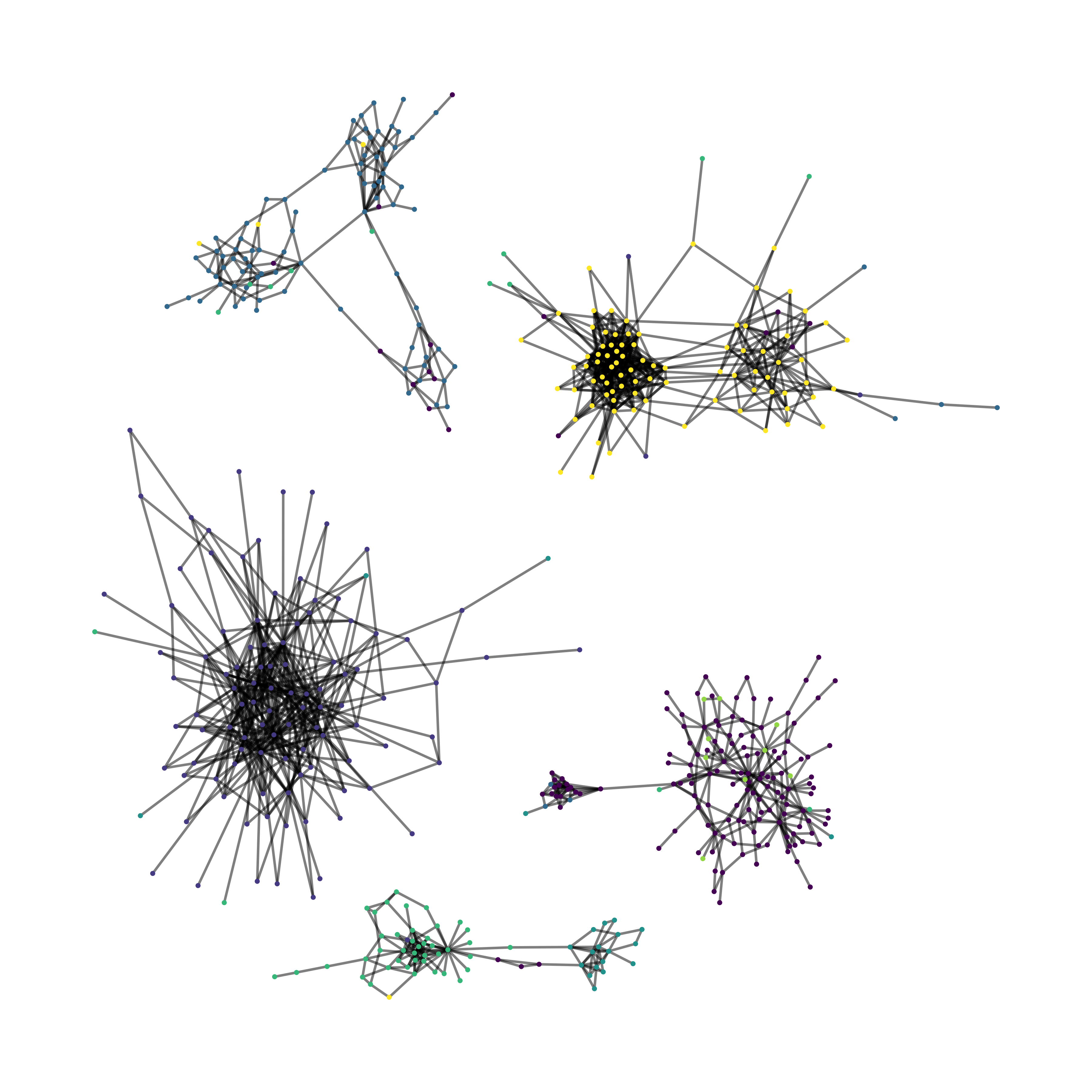}
         \caption{HiGGs Sampled}
         \label{fig:cora_higgs_comm}
     \end{subfigure}
     % \hfill
     \centering
     \begin{subfigure}[b]{0.48\textwidth}
         \centering
         \includegraphics[width=\textwidth]{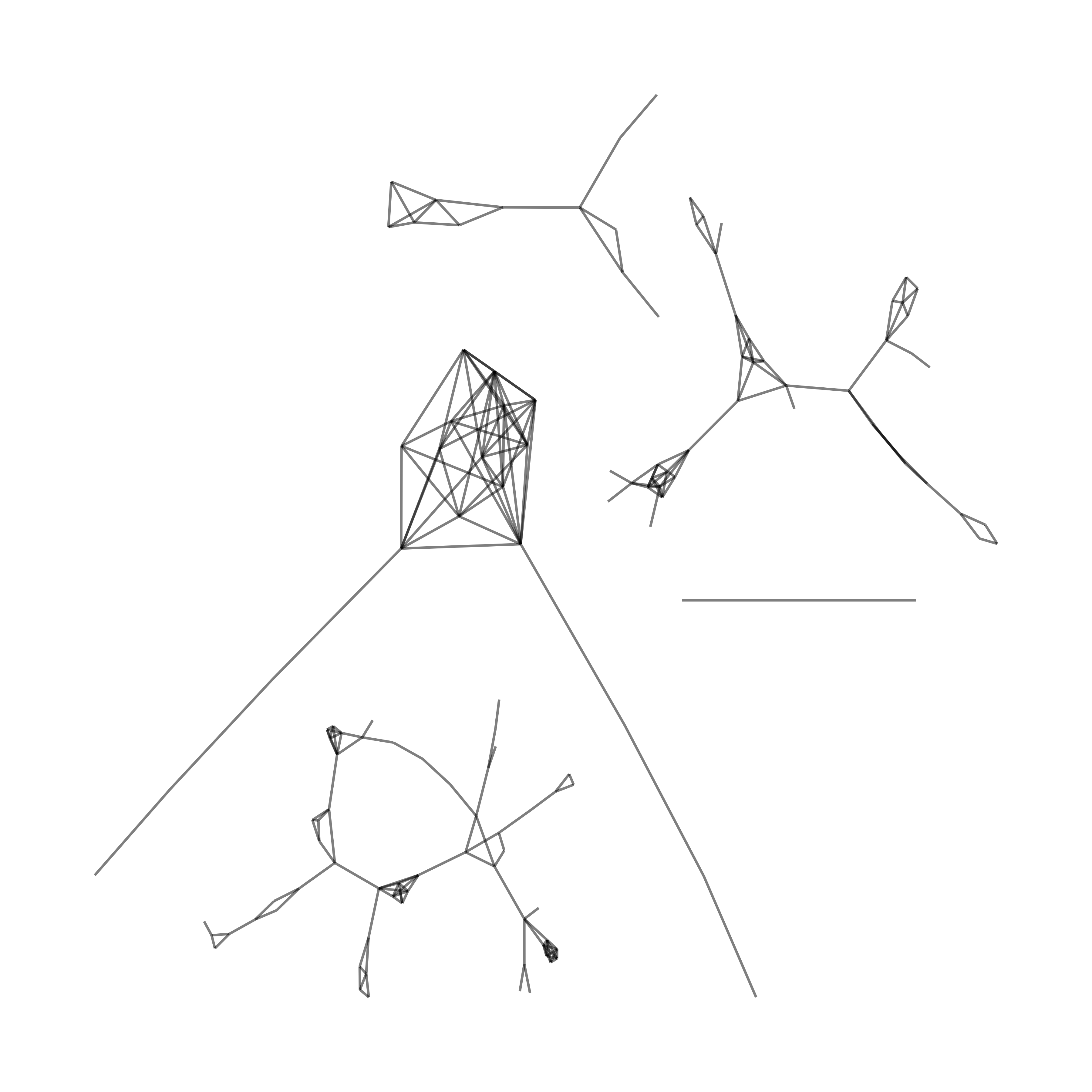}
         \caption{BTER Sampled}
         \label{fig:cora_bter_comm}
     \end{subfigure}
        \caption{Communities sampled with Louvain detection from CORA graph with the same from \method and BTER generated counterparts, with node-level QQ plots. Layouts are through SFDP and GraphViz.}
        \label{fig:cora_synth_real_comm}
\end{figure}

\begin{figure}[ht]
    \centering
    \begin{subfigure}[b]{\textwidth}
        \centering
        \includegraphics[width=\textwidth]{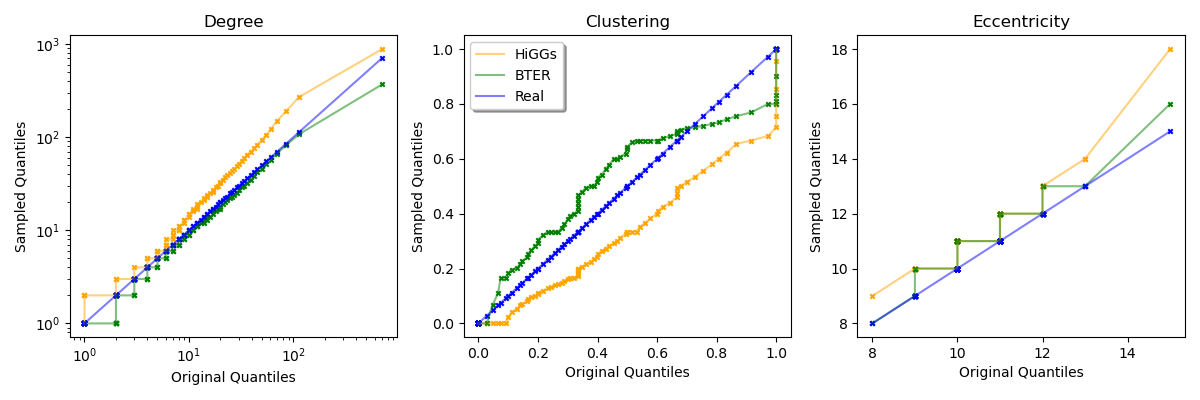}
        \caption{Whole-graph QQ plots}
        \label{fig:fb_qq}
    \end{subfigure}
     \centering
     \begin{subfigure}[b]{0.48\textwidth}
         \centering
         \includegraphics[width=\textwidth]{images/Real_fb.png}
         \caption{Original}
         \label{fig:fb_real}
     \end{subfigure}
     \hfill
     \begin{subfigure}[b]{0.48\textwidth}
         \centering
         \includegraphics[width=\textwidth]{images/HiGGs_fb.png}
         \caption{HiGGs Sampled}
         \label{fig:fb_higgs}
     \end{subfigure}
     % \hfill
     \centering
     \begin{subfigure}[b]{0.48\textwidth}
         \centering
         \includegraphics[width=\textwidth]{images/BTER_fb.png}
         \caption{BTER Sampled}
         \label{fig:fb_bter}
     \end{subfigure}
        \caption{The Facebook Page-Page graph with sampled counterparts from \method and BTER, with node-level QQ plots. Layouts are through SFDP and GraphViz.}
        \label{fig:fb_synth_real}
\end{figure}

\begin{figure}[ht]
    \centering
    \begin{subfigure}[b]{\textwidth}
        \centering
        \includegraphics[width=\textwidth]{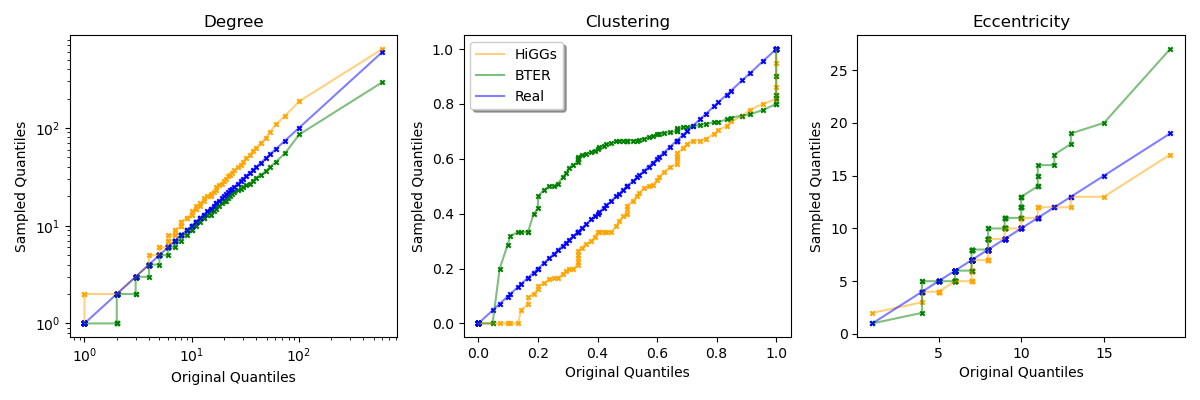}
        \caption{Resampled community QQ plots}
        \label{fig:fb_qq_comm}
    \end{subfigure}
     \centering
     \begin{subfigure}[b]{0.48\textwidth}
         \centering
         \includegraphics[width=\textwidth]{images/FB_real_comms.png}
         \caption{Original}
         \label{fig:fb_real_comm}
     \end{subfigure}
     \hfill
     \begin{subfigure}[b]{0.48\textwidth}
         \centering
         \includegraphics[width=\textwidth]{images/FB_HiGGs_comms.png}
         \caption{HiGGs Sampled}
         \label{fig:fb_higgs_comm}
     \end{subfigure}
     % \hfill
     \centering
     \begin{subfigure}[b]{0.48\textwidth}
         \centering
         \includegraphics[width=\textwidth]{images/FB_bter_comms.png}
         \caption{BTER Sampled}
         \label{fig:fb_bter_comm}
     \end{subfigure}
        \caption{Communities sampled with Louvain detection from Facebook Page-Page graph with the same from \method and BTER generated counterparts, with node-level QQ plots. Layouts are through SFDP and GraphViz.}
        \label{fig:fb_synth_real_comm}
\end{figure}

\subsection{Assumptions and Threats to Validity}

A limitation in this work is that we produce only a baseline implementation \method using one set of models.
We do it for simplicity, and to keep the scope of the work reasonable, but does mean that we cannot draw comparisons between possible implementations.
We also consider an ablation study, with permutations of different candidate models for each stage, beyond the scope of this work.
A different partitioning algorithm during training data creation could also have beneficial results, particularly one that aims to produce partitions of a near-uniform size, which would allow for smaller models as the maximum size of training sample graphs decreases.

Further, with different component generative models the upper limit on the size of generated graphs could be far higher, as with \method the possible size of generated graphs scales $|V_{max}|^2$ \footnotesize{with $|V_{max}|$ the lowest upper limit on graph size for the models used}.
If one were to implement \method using GRAN from \citet{Liao2019EfficientNetworks} with a suitably efficient edge-sampling model, the feasible number of nodes in generated graphs could extend \textit{above a million nodes}.% with its demonstrated capacity to produce graphs $|V_{max}| > 1000$.

One core assumption of our \method implementation here is that the edges between $h_1$ graph pairs can be reasonably sampled without representation of where other $h_1$ graphs have been connected.
This makes our implementation trivially paralellisable, as a large set of separate graph or edge sampling tasks, but does make it somewhat greedy.
Feasibly passing of information from prior edge samplings, perhaps through node embeddings or similar, may improve performance --- a possible future direction.
Additionally in this work, we do not apply \method to any geometric or quasi-geometric graphs such as molecules or road networks.
The motivation behind this was again to limit the scope of this work, as these graphs have fragile topological features such as near-planarity or strict connection rules as in chemistry.
Such graphs would then need a different edge sampling process, which considers prior connections. We leave this for future work.

% \section{Section 1}

% \subsection{Subsection 1}

% \appendix

% \section{Appendix}

% \bibliographystyle{}
% \bibliographystyle{IEEEtranN}

\end{document}